\RequirePackage{fix-cm}
\documentclass[smallextended]{svjour3}       
\smartqed  
\usepackage[pdftex]{graphicx}
\usepackage[utf8]{inputenc} 
\usepackage[T1]{fontenc}    
\usepackage{hyperref}       
\usepackage{url}            
\usepackage{booktabs}       
\usepackage{amsfonts}       
\usepackage{nicefrac}       
\usepackage{microtype}      
\usepackage{subfigure}
\usepackage{bm}
\usepackage{color}
\usepackage{times}
\usepackage[export]{adjustbox}
\usepackage{soul}

\usepackage{algorithm}
\usepackage{algorithmic}
\usepackage{amssymb}
\usepackage{amsmath}
\usepackage[skip=2pt]{caption}

\newtheorem{assumption}{Assumption}
\usepackage[round]{natbib}

\usepackage[symbol]{footmisc}

\begin{document}

\title{Scaling up Stochastic Gradient Descent for Non-convex Optimisation}


\author{Saad~Mohamad \and Hamad~Alamri \and Abdelhamid~Bouchachia\footnote{Corresponding author. \email{abouchachia@bournemouth.ac.uk}} 
}

\authorrunning{S. Mohamad, H. Alamri, A. Bouchachia} 
\institute{S. Mohamad and A. Bouchachia \at
              Department of Computing, Bournemouth University, Poole, UK
\and H. Alamri \at WMG, Warwick University, Coventry, UK
}

\date{Received: date / Accepted: date}

\maketitle

\begin{abstract}
Stochastic gradient descent (SGD) is a widely adopted iterative method for optimizing  differentiable objective functions. In this paper, we propose and discuss a novel approach to scale up SGD in applications involving non-convex functions and large datasets. We address the bottleneck problem arising when using both shared and distributed memory. Typically, the former is bounded by limited computation resources and bandwidth whereas the latter suffers from communication overheads. We propose a  unified distributed and parallel implementation of SGD (named DPSGD) that relies on both asynchronous distribution and lock-free parallelism. By combining two strategies into a unified framework, DPSGD is able to strike a  better trade-off between local computation and communication. The convergence properties of DPSGD are studied for non-convex problems such as those arising in statistical modelling and machine learning. Our theoretical analysis shows that DPSGD leads to speed-up with respect to the number of cores and number of workers while guaranteeing an asymptotic convergence rate of $O(1/\sqrt{T})$ given that the number of cores is bounded by $T^{1/4}$ and the number of workers is bounded by $T^{1/2}$ where $T$ is the number of iterations. The potential gains that can be achieved by DPSGD are demonstrated empirically on a stochastic variational inference problem (Latent Dirichlet Allocation) and on a deep reinforcement learning (DRL) problem (advantage actor critic - A2C) resulting in two algorithms: DPSVI and HSA2C.  Empirical results validate our theoretical findings. Comparative studies are conducted to show the performance of the proposed DPSGD against the state-of-the-art DRL algorithms.

\end{abstract}

\section{Introduction}\label{intro}

Stochastic gradient descent (SGD) is a general iterative algorithm for solving large-scale optimisation problems such as minimising a differentiable objective function $f(\boldsymbol v)$ parameterised in $\boldsymbol v \in \mathcal{V}$: 
\begin{equation}\label{intro_1}
\min_{\boldsymbol v} f(\boldsymbol v) 
\end{equation}
In several statistical models and machine learning  problems, $f(\boldsymbol v)=\frac{1}{n}\sum_{i=1} ^{n}f_i(\boldsymbol v)$ is the empirical average loss where each $f_i(\boldsymbol v)$ indicates that the function has been evaluated at a data instance $\boldsymbol x_i$. SGD updates $\boldsymbol v$ using the gradients computed on some (or single) data points. In this work, we are interested in problems involving non-convex objective functions such as variational inference~\citep{blei2017variational,hoffman2013stochastic,wainwright2008graphical,jordan1999introduction} and artificial neural networks ~\citep{chilimbi2014project,dean2012large,li2014scaling,xing2015petuum,zhang2015deep} amongst many others. Non-convex problems abound in ML and are often characterised by a very large number of parameters (e.g., deep neural nets), which hinders their optimisation. This challenge is often compounded by the sheer size of the training datasets, which can be in order of millions of data points.  As the size of the available data increases, it becomes more essential to boost SGD scalability  by distributing and parallelising its sequential computation. The need for scalable optimisation algorithms is shared across application domains. 

Many studies have proposed to scale-up SGD by distributing the computation over different computing units taking advantage of the advances in hardware technology. The main existing paradigms exploit either \textit{shared memory} or \textit{distributed memory} architectures. While shared memory is usually used to run an algorithm in parallel on a single multi-core machine~\citep{recht2011hogwild,zhao2017lock,lian2015asynchronous,huo2016asynchronous}, distributed memory, on the other hand, is used to distribute the algorithm on multiple machines~\citep{agarwal2011distributed,lian2015asynchronous,zinkevich2010parallelized,langford2009slow}. Distributed SGD (DSGD) is appropriate for very large-scale problems where data can be distributed over massive (theoretically unlimited) number of machines each with its own computational resources and I/O bandwidth. However DSGD efficiency is bounded by the communication latency across machines. Parallel SGD (PSGD) takes advantage of the multiple and fast processing units within a single machine with higher bandwidth communication; however, the computational resources and I/O bandwidth are limited.

In this paper, we set out to explore the potential gains that can be achieved by leveraging the advantages of both the distributed and parallel paradigms in a unified approach. The proposed algorithm, DPSGD, curbs the communication cost by updating a local copy of the parameter vector being optimised, $\boldsymbol v$, multiple times during which each machine performs parallel computation. The distributed computation of DPSGD among multiple machines is carried out in an \textit{asynchronous} fashion whereby workers compute their local updates independently~\citep{lian2015asynchronous}. A master aggregates these updates to amend the global parameters. The parallel implementation of DPSGD on each local machine is {\it lock-free} whereby multiple cores are allowed equal access to the shared memory to read and update the variables without locking~\citep{zhao2017lock} (i.e., they can read and write the shared memory simultaneously). We provide a theoretical analysis of the convergence rate of DPSGD for non-convex optimisation problems and prove that linear speed-up with respect to the number of cores and workers is achievable while they are bounded by $T^{1/4}$ and $T^{1/2}$, respectively, where $T$ is the total number of iterations. Furthermore, we empirically validate these results by developing two inferential algorithms relying on DPSGD. The first one is an asynchronous lock-free stochastic variational inference algorithm (DPSVI) that can be deployed on a wide family of Bayesian models (see the Appendix); its potential is demonstrated here on a Latent Dirichlet Allocation problem. The second one is DPSGD-based Deep Reinforcement Learning (DRL) that can be used to scale up the training of DRL networks for multiple tasks (see the Appendix). 

The rest of the paper is organised as follows. Section~\ref{related} presents the related work. Section~\ref{Sec3} presents the proposed algorithm and the theoretical study. In Section~\ref{sec4}, we carry out experiments and discuss the empirical results. Finally, Section~\ref{sec7} draws some conclusions and suggests future work. The appendices present the proofs, the asynchronous distributed lock-free parallel SVI and the highly-scalable actor critic algorithm.

\section{Related work}\label{related} 

We divide this section into two parts. In the first part, we discuss the relevant literature to distributed and parallel SGD focusing on the theoretical aspects~\citep{lian2015asynchronous,zhao2017lock,fang2017parallel,huo2017asynchronous,bottou2010large,niu2011hogwild,tsitsiklis1986distributed,elgabli2020q,wang2019matcha,recht2011hogwild,leblond2017asaga,dean2012large,zhou2017convergence,yu2019parallel,lin2018don,stich2018local}. To keep the paper centred and due to space limitation, only SGD-based methods for non-convex problems are covered. The second part covers related work focusing on the implementation/application of the distributed and parallel algorithms~\citep{hoffman2013stochastic,smohamad,neiswanger2015embarrassingly,dean2012large,paine2013gpu,ruder2016overview,li2014scaling,abadi2016tensorflow,paszke2017automatic,babaeizadeh2016reinforcement,mnih2016asynchronous,clemente2017efficient,horgan2018distributed,espeholt2018impala,nair2015massively,adamski2018distributed}.

\subsection{Theoretical aspects}
A handful of SGD-based methods have been proposed recently for large-scale non-convex optimisation problems~\citep{de2015taming,lian2015asynchronous,zhao2017lock,fang2017parallel,huo2017asynchronous}, which embrace either a distributed or parallel paradigm. HOGWILD  presented by~\cite{niu2011hogwild} proposes several asynchronous parallel SGD variants with locks and lock-free shared memory. Theoretical convergence analysis for convex objectives presented in that study was inspiring and adopted for most of the recent literature on asynchronous parallel optimisation algorithms. Similarly~\cite{leblond2017asaga,de2015taming} provided convergence analysis for SGD asynchronous lock-free parallel optimisation with shared memory for convex objectives where they provide convergence analysis with relaxed assumptions on the sparsity of the problem.~\cite{de2015taming} also analysed the HOGWILD convergence for non-convex objectives. 
Asynchronous distributed and lock-free parallel SGD algorithms for non-convex objectives have also been studied in~\citep{lian2015asynchronous} showing that linear speed-up with respect to the number of workers is achievable when bounded by $O(\sqrt{T})$. Improved versions using variance reduction techniques have recently been proposed in~\citep{huo2017asynchronous,fang2017parallel} to accelerate the convergence rate with a linear rate being achieved instead of the sub-linear one of SGD. Although these and other algorithmic implementations are lock-free~\citep{lian2015asynchronous,huo2017asynchronous,fang2017parallel,de2015taming}, the theoretical analysis of the convergence was based on the assumption that no over-writing happens. Hence, write-lock or atomic operation for the memory are needed to prove the convergence. In contrast,~\cite{zhao2017lock} proposed a completely parallel lock-free implementation and analysis.

Different implementations exploiting both  parallelism with shared memory and distributed computation across multiple machines have been proposed~\citep{dean2012large,zhou2017convergence,yu2019parallel,lin2018don,stich2018local}. Except for ~\citep{stich2018local,dean2012large}, these methods adopted synchronous SGD implementation where~\cite{dean2012large,lin2018don} focused on the implementation aspects, providing extensive empirical study on deep learning models. While the implementation ideas are very similar to ours, we consider lock-free local parallelism with asynchronous distribution and we provide theoretical analysis. We also evaluate our approach on different ML problems i.e.,  SVI and DRL (to be discussed in the next part). Instead of using a parameter server, the local learners in~\citep{zhou2017convergence} compute the average of their copies of parameters at regular intervals through global reduction. Communication overhead is controlled by introducing a communication interval parameter into the algorithm. However, the provided theoretical analysis in~\citep{zhou2017convergence} does not establish a speedup and synchronisation is required for global reduction. Authors in~\citep{yu2019parallel} provide theoretical study for the model averaging introduced  in~\citep{zhou2017convergence} showing that linear speedup of local SGD on non-convex
objectives can be achieved as long the averaging interval is carefully controlled. Similar study for convex problems with asynchronous worker communication by~\cite{stich2018local} shows linear speedup in the number of workers and the mini-batch size with reduced communication. The multiple steps local SGD update by~\cite{stich2018local} aimed at reducing the communication overhead is similar to our proposed algorithm. Nonetheless, we adopt local lock-free parallelism and asynchronous distribution with parameter server scheme instead of model averaging. Finally, we point out that although we focus on SGD first-order method~\citep{bottou2010large,niu2011hogwild,tsitsiklis1986distributed,elgabli2020q,wang2019matcha}, our study can be extended to second-order methods~\citep{shamir2014communication,jahani2020efficient,ba2016distributed,crane2019dingo,jahani2020scaling} and variance reduction methods~\citep{huo2017asynchronous,fang2017parallel}, where the high noise of our local multiple-steps update can be reduced contributing to further speed-up. We leave this for future work.

\subsection{Implementation aspects}
The first effort to scale-up variational inference is described in ~\citep{hoffman2013stochastic} where gradient descent updates are replaced with SGD updates. Inspired by this work,~\cite{smohamad} replaces the SGD updates with asynchronous distributed SGD ones. This was achieved by computing the SVI stochastic gradient on each worker based on few (mini-batched or single) data points acquired from distributed sources. The update steps are then aggregated to form the global update. In \citep{neiswanger2015embarrassingly}, the strategy consists of distributing the entire dataset across workers and letting each one of them perform VI updates in parallel. This requires that, at each iteration, the workers must be synchronised to combine their parameters. However, this synchronisation requirement limits the scalability so the maximum speed achievable is bounded by the slowest worker. Approaches for scaling up VI that rely on Bayesian filtering techniques have been reviewed in ~\citep{smohamad}.

Asynchronous SGD (ASYSG)~\citep{lian2015asynchronous},  an implementation of SGD that distributes the training over multiple workers, has been adopted by DistBelief~\citep{dean2012large} (a parameter server-based algorithm for training neural networks) and Project Adam~\citep{chilimbi2014project} (another DL framework for training neural networks).  \cite{paine2013gpu} showed that ASYSG can achieve noticeable speedups on small GPU clusters. Other similar work~\citep{ruder2016overview,li2014scaling} have also employed ASYSG to scale up deep neural networks. The two most popular and recent DL frameworks TensorFlow~\citep{abadi2016tensorflow} and Pytorch \citep{paszke2017automatic} have embraced the Hogwild~\citep{recht2011hogwild,zhao2017lock} and ASYSG~\citep{lian2015asynchronous} implementations to scale-up DL problems\footnote{See, for instance, https://github.com/pytorch/examples and https://github.com/tmulc18/Distributed-TensorFlow-Guide}.

Distributed and parallel SGD have also been employed in deep reinforcement learning (DRL)~\citep{babaeizadeh2016reinforcement,mnih2016asynchronous,clemente2017efficient,horgan2018distributed,espeholt2018impala,nair2015massively,adamski2018distributed}. In~\citep{babaeizadeh2016reinforcement}, a hybrid CPU/GPU version of the Asynchronous Advantage ActorCritic (A3C) algorithm~\citep{mnih2016asynchronous} was introduced. The study focused on mitigating the severe under-utilisation of the GPU computational resources in DRL caused by its sequential nature of data generation. Unlike~\citep{mnih2016asynchronous}, the agents in RL do not compute the gradients themselves. Instead, they send data to central learners that update the network on the GPU accordingly. However, as the number of core increases, the central GPU learner becomes unable to cope with the data. Furthermore, large amount of data requires large storage capacity. Also the internal communication overhead can affect the speed-up when the bandwidth reaches its ceiling. We note that a similar way for paralleling DRL is proposed by~\cite{clemente2017efficient}. Similarly, \cite{horgan2018distributed} propose to generate data in parallel using multi-cores CPU's where experiences are accumulated in a shared experience replay memory. Along the same trend, \cite{espeholt2018impala} proposed to accumulate data by distributed actors and communicate it to the centralised learner where the computation is done.  The architecture of these studies~\citep{horgan2018distributed,espeholt2018impala} allows the distribution of the generation and selection of data instead of distributing locally computed gradients as in~\citep{nair2015massively}. Hence, it requires sending large size information over the network in case of large size batch of data making the communication more demanding.  Furthermore, the central learner has to perform  most of the computation limiting the scalability.

The work in ~\citep{adamski2018distributed} is the most similar to ours, where SGD based hybrid distributed-parallel actor critic is studied. The parallel algorithm of~\citep{mnih2016asynchronous} is combined with parameter server architecture of~\citep{nair2015massively} to allow parallel distributed implementation of A3C on a computer cluster with multi-core nodes. Each node applies the algorithm in~\citep{babaeizadeh2016reinforcement} to queue data in batches, which are used to compute local gradients. These gradients are then gathered from all workers, averaged and applied to update the global parameters. To reduce the communication overhead, authors carried out careful reexamination of Adam optimiser’s hyper-parameters allowing large-size batches to be used. Detailed discussion of these methods and comparison  to our implementation is provided in the appendix. 

\section{The DPSGD algorithm and its properties}\label{Sec3}

Before delving into the details of the proposed algorithm, we introduce the list of symbols in Tab.~\ref{tab001sym} that are used in the rest of the text.

\begin{table}[htb]

\caption{\textbf{List of symbols. }}

\centering
\begin{tabular}{lp{10cm}}
\toprule
Variable & Description \\
\toprule
$\boldsymbol v$ & The parameter vector to be optimised, called the {\it global parameter}. It is maintained and updated by the master machine\\

$\boldsymbol u$ & Copy of the global parameter maintained and is updated by the workers\\

$\boldsymbol{\hat u}$ & Copy of the local parameter $\boldsymbol u$ and is stored in the worker's shared memory\\

$||\boldsymbol x||$& The Euclidean norm of vector $\boldsymbol x$\\

$f_i(.)$ & The objective function defined on the $i^{th}$ instance\\

$\nabla f(\boldsymbol{v})$  & The gradient vector of $f(\boldsymbol{v})$\\

$t$ & The global unique iterate used in the synthetic sequence  - not required by the algorithm \\

$b$ & The local unique iterate used in the synthetic sequence - not required by the algorithm \\

$m$ & Index referring to the update vector computed by the worker $n_m$ and collected by the master\\ 

$M$ & The master batch size which is the number of local updates coming from the workers for each global update by the master.\\

$p$ & The total number of threads (cores)\\

$nW$ & The total number of workers (nodes)\\

$S$ and $P$ & Binary diagonal matrices used to denote whether local over-writing happens (see Sec.~\ref{senssequ}). \\
  
\bottomrule
\end{tabular}
\label{tab001sym}
\end{table}

\subsection{Overview of the algorithm}

The proposed DPSGD algorithms assumes a \textit{star-shaped} computer network architecture: a master maintains the global parameter $\boldsymbol v$ (Alg.~\ref{alg21}) and the other machines act as workers which independently and simultaneously compute local parameters $\boldsymbol{u}$ (Alg.~\ref{alg22}). The workers communicate only with the master in order to access the state of the global parameter (line 3 in Alg. \ref{alg22}) and provide the master with their local updates (computed based on local parameters) (line 10 in Alg. ~\ref{alg22}). Each worker is assumed to be a multi-core machine, and the local parameter are obtained by running a lock-free parallel SGD (see Alg. ~\ref{alg22}). This is achieved by allowing all cores equal access to the shared memory to read and update at any time with no restriction at all~\citep{zhao2017lock}. The master aggregates $M$ predefined amounts of local updates coming from the workers (line 3 in Alg.~\ref{alg21}), and then computes its global parameter. The update step is performed as an atomic operation such that the workers are locked out and cannot read the global parameter during this step (see Alg.~\ref{alg21}). 

\begin{algorithm}[t]
   \caption{DPSGD-Master: Updates performed at the master machine}\label{alg21}
\begin{algorithmic}[1]
       \STATE \textbf{initialise:} number of iteration $T$, global variable $\boldsymbol v$, global learning rate $\{\rho_t\}_{t=0,...,T-1}$ 
       \FOR{$t=0,1,2,...T-1$}
        \STATE Collect $M$ updating vectors $\boldsymbol w_1,...,\boldsymbol w_M$ from the workers.
      \STATE Update the current estimate of the global parameter $\boldsymbol v \leftarrow \boldsymbol v+\rho_{t}\sum_m \boldsymbol w_m$
      \STATE $t\leftarrow t+1$
     \ENDFOR
\end{algorithmic}
\end{algorithm}
\begin{algorithm}[t]
   \caption{DPSGD-Worker: Updates performed at each worker machine}\label{alg22}
\begin{algorithmic}[1]
  \STATE \textbf{initialise:} number of iterations of per-worker loop $B$, learning rate $\eta$, number of threads $p$ 
   \WHILE{(MasterIsRun)}
        \STATE Pull a global parameter $\boldsymbol v$ from the master and put it in the shared memory.
        \STATE Fork p threads
        \FOR{c=0 to B-1}
            \STATE Read current values of $\boldsymbol u$, denoted as $\boldsymbol{\hat u}$, from the shared memory.
        \STATE Randomly pick i from $\{1,...n\}$ and compute the gradient $\nabla f_i(\boldsymbol{\hat u})$.
        \STATE $\boldsymbol{ u}\leftarrow \boldsymbol{ u} -\eta\nabla f_i(\boldsymbol{\hat u})$
     \ENDFOR
     \STATE Push the update vector $\boldsymbol{u}-\boldsymbol{v}$ from the shared memory to the master
     \ENDWHILE
\end{algorithmic}
\end{algorithm}

Note that the local distributed computations are done in an asynchronous style, i.e., DPSGD does not lock the workers until the master starts updating the global parameter. That is, the workers might compute some of the stochastic gradients based on early values of the global parameter. Similarly, the lock-free parallel implies that local parameter can be updated by other cores in the time after being read and before being used for the update computation. Given this non-synchronisation among workers and among cores, the results of parameter update seem to be totally disordered, which makes the convergence analysis very difficult.

Following~\cite{zhao2016fast}, we introduce a synthetic process to generate the final value of local and global parameters after all threads, workers have completed their updates as shown in Alg.~\ref{alg21} and~\ref{alg22}. That is, we generate a sequence of synthetic values of $\boldsymbol v$ and $\boldsymbol{u}$ with some order to get the final value of $\boldsymbol{v}$. These synthetic values are used for DPSGD convergence proof. The synthetic generation process is explained in the following section. 

\subsection{Synthetic Process}\label{senssequ}

Let $t$ be the global unique iterate attached to the loop in Alg.~\ref{alg21}; $b$  is the local unique iterate attached to the inner loop in Alg.~\ref{alg22} and $m$ is an index referring to the update vector computed by a worker $n_m\in \{1, .., nW\}$. If we omit the outer loop of Alg.~\ref{alg22}, the key steps in Alg.~\ref{alg22} are those related to the writing (updating) or reading the local parameter. 

\subsubsection{Local Synthetic Write (Update) Sequence}  
As in~\citep{zhao2016fast}, we assume all threads will update the elements in $\textbf{u}$ in the order from $1$ to $\tilde B$, where $\tilde B= B*p$ with $p$ is the number of threads. Thus, $\{u_1,…u_{{\tilde B}-1}\}$ is the  synthetics sequence which may never occur in the shared memory. However, it is employed to obtain the  final value $u_{\tilde B}$ after all threads have completed their updates in the inner-loop of Alg.~\ref{alg22}. In other terms, this  ordered synthetic update sequence generating  the same final value as that of the disordered lock-free update process. At iterate $b$, the synthetic update done by a thread can be written as follows: 
\begin{equation}\label{equ_ne_1}
\boldsymbol{ u}_{b}=\boldsymbol{ u}_{0}-\sum_{j=0}^{b-1}\eta S_j \nabla f_{i_{j}}(\boldsymbol{\hat u}_{j})
 \end{equation}
where $S_j$ is a diagonal matrix whose entries are $0$ or $1$, determining which dimensions of the parameter vector $u_b$ have been successfully updated by the $j_{th}$ gradient  computed on the shared local parameter $\hat{u}_j$ . That is $S_j(k,k)=0$ if dimension $k$ is over-written by another thread and $S_j(k,k)=1$ if dimension $k$ is successfully updated by $\nabla f_{i_{j}}(\boldsymbol{\hat u}_{j})$  without over-writing.  Equation~\ref{equ_ne_1} can be rearranged in an iterative form as:
\begin{equation}\label{equ_ne_2}
\boldsymbol{ u}_{b+1}=\boldsymbol{ u}_{b}-\eta S_{b} \nabla f_{i_{b}}(\boldsymbol{\hat u}_{b})
 \end{equation}
Including the outer loop and the global update in Alg.~\ref{alg21}, we define the synthetic sequence $\{\boldsymbol u_{t,m,b}\}$ equivalent to the updates for the $b^{th}$ per-worker loop of the $m^{th}$ update vector associated with the $t^{th}$ master loop:
\begin{align}\label{equ_2_1}
\nonumber&\text{\text{Algorithm~\ref{alg22}, line 3 refers to: }}
\nonumber \boldsymbol u_{t,m,0}=\boldsymbol v_{t-1}\\
\nonumber &\text{\text{Algorithm~\ref{alg22}, line 8 refers to: }}
\nonumber \boldsymbol u_{t,m,b+1}=\boldsymbol u_{t,m,b}-\eta S_{t+\tau_{t,m},m,b}\nabla f_{i_{t+\tau_{t,m},m,b}}(\boldsymbol{\hat u}_{t,m,b})\\
 &\text{\text{Algorithm~\ref{alg21}, line 4 refers to: }}
   \boldsymbol v_{t}=\boldsymbol v_{t-1}+\rho_{t-1}(\sum_{m=1}^{M}\boldsymbol u_{t-\tau_{t,m},m,\tilde B}-\boldsymbol v_{t-1-\tau_{t,m}})
\end{align}
 where $\tau_{t,m}$ is the delay of the $m^{th}$ global update for the $t^{th}$ iteration caused by the asynchronous distribution. To compute $\nabla f_{i_{t+\tau_{t,m},m,b}}(\boldsymbol{\hat u}_{t,m,b})$, $\boldsymbol{\hat u}_{t,m,b}$ is read from the shared memory by a thread.

\subsubsection{Local memory read} As denoted earlier, $\hat{u}_b$ is the local parameter read from the shared memory which is used to compute $\nabla f_{i_{b}}(\boldsymbol{\hat u}_{b})$ by a thread. Using the synthetic sequence $\{u_1,..u_{\tilde B-1}\}$,  $\hat{u}_b$ can be written as:
\begin{equation}\label{equ_ne_3}
 \hat{u}_b  =u_{a( b)}+\sum_{j=a(b)}^{b-1}P_{b,j-a(b)}\nabla f_{i_j}(\hat{u}_j) 
 \end{equation}
where $a(b) <b$ is the step in the inner-loop whose updates have been completely written in the shared memory.  $P_{b,j-a(b)}$ are diagonal matrices whose diagonal entries are $0$ or $1$. $\sum_{j=a(b)}^{b-1}P_{b,j-a(b)}\nabla f_{i_j}(\hat{u}_j)$ determine what dimensions of the new gradient updates, $\nabla f_{i_j}(\hat{u}_j)$, from time $a(b)$ to $b-1$ have been added to $u_{a(b)}$  to obtain $\hat{u}_b$. That is,  $\hat{u}_b$ may read some dimensions of new gradients between time $a(b)$ to $b-1$ including those which might have been over-written by some other threads. Including the outer loop and the global update in Alg.~\ref{alg21}, the local read becomes:
  \begin{equation}\label{equ_2_2}
    \boldsymbol{\hat u}_{t,m,b}=\boldsymbol{ u}_{t,m,a(b)}-\eta\sum_{j=a(b)}^{b-1}P^{t+\tau_{t,m},m}_{b,j-a(b)}\nabla f_{i_{t+\tau_{t,m},m,j}}(\boldsymbol{\hat u}_{t,m,j})
   \end{equation}
 The partial updates of the remaining steps between  $a(b)$ and $b-1$ are now defined by $\{P^{t,m}_{b,j-a(b)}\}_{a(b)}^{b-1}$.

\subsection{Convergence analysis} 

Using the synthetic sequence, we develop the theoretical results of DPSGD showing that under some assumptions, we can guarantee linear speed-up with respect to the number of cores (threads) and number of nodes (workers). Before presenting the studies, we introduce and explain the require assumptions:
   
  \begin{assumption}
  The function $f(.)$ is smooth, that is to say, the gradient of $f(.)$ is \textit{Lipschitzian}: there exists a constant $L>0$, $\forall\boldsymbol x, \boldsymbol y$, 
  \begin{equation*}
  ||\nabla f(\boldsymbol x)-\nabla f(\boldsymbol y)||\leq L||\boldsymbol x-\boldsymbol y||
  \end{equation*}
   or equivalently,
    \begin{equation*}
f(\boldsymbol y)\leq f(\boldsymbol x)+ \nabla f(\boldsymbol x)^T(\boldsymbol y-\boldsymbol x)+ \frac{L}{2}||\boldsymbol y-\boldsymbol x||^2.
  \end{equation*}
 \end{assumption}
  
    \begin{assumption}
    The per-dimension over-writing defined by $S_{t,m,b}$ is a random variate, independent of $i_{t,m,j}$\footnote{Note that $i$ can be a set of indices for a per-worker mini-batch. In this paper, $i$ refers to a single index for simplicity}.
    \end{assumption}
    
     This assumption is reasonable since $S_{t,m,b}$  is affected by the hardware, while $i_{t,m,j}$ is independent thereof.
    \begin{assumption}
The conditional expectation of the random matrix $S_{t,m,b}$ on $\boldsymbol{u}_{t,m,b}$ and $\boldsymbol{\hat u}_{t,m,b}$ is a strictly positive definite matrix, i.e., $\mathbb{E}[S_{t,m,b}| \boldsymbol{u}_{t,m,b}, \boldsymbol{\hat u}_{t,m,b}] =S\succ 0$ with the minimum eigenvalue $\alpha>0$. 
\end{assumption}
\begin{assumption}
The gradients are unbiased and bounded: $\nabla f(\boldsymbol x)=\mathbb{E}_i[\nabla f_i(\boldsymbol x)]$ and $||\nabla f_i(\boldsymbol x)||\leq V$, $\forall i \in\{1,...n\}$. 
\end{assumption}
Then, it follows that the variance of the stochastic gradient is bounded. $\mathbb{E}_i[||\nabla f_i(\boldsymbol x)-\nabla f(\boldsymbol x)||^2]\leq \sigma^2$, $\forall \boldsymbol x$, where $\sigma^2=V^2-||\nabla f(x)||$
\begin{assumption}
Delays between old local stochastic gradients and the new ones in the shared memory are bounded: $0\leq b-a(b)\leq D$ and the delays between stale distributed update vectors and the current ones are bounded $0\leq \max_{t,m}\tau_{t,m}\leq D'$
\end{assumption}
\begin{assumption} 
All random variables in $\{i_{t,m,j}\}_{\forall t, \forall m, \forall j}$ are independent of each other. 
\end{assumption}  
Note that we are aware that this independence assumption is not fully accurate due to the potential dependency between selected data samples for computing gradients at the same shared parameters.  For example, samples with fast computation of gradients for the same shared variable leads to more frequent selection of these samples as they likely to finish their gradient computation while the shared memory has not been overwritten. Hence, the selected samples can be correlated. This can also affect the independence assumption between the overwriting matrix and the selected sample (Assumption 2). However, we follow  existing studies~\cite{zhao2016fast,zhao2017lock,lian2015asynchronous,reddi2015variance,duchi2015asynchronous,de2015taming,lian2018asynchronous,hsieh2015passcode}, assuming DPSGD maintains the required conditions for independence via Assumptions 2 and 6.

We are now ready to state the following convergence rate for any non-convex objective:

\begin{theorem}
If Assumptions (1) to (6) hold and the following inequalities are true:
\begin{equation}\label{proof_133}
 M^2\tilde B^2\eta^2L^2\rho_{t-1}D'\sum_{n=1}^{D'}\rho_{t+n}\leq 1
\end{equation}
\begin{equation}\label{proof_66}
\frac{1}{1-\eta-\frac{9\eta(D+1)L^2(\mu^{D+1}-1)}{\mu-1}}\leq \mu 
\end{equation}
\text{then, we can obtain the following results:}
\begin{align}
\nonumber &\frac{1}{\sum_{t=1}^{T}\rho_{t-1}}\sum_{t=1}^{T}\rho_{t-1} \mathop{\mathbb{E}}[||\nabla f(\boldsymbol v_{t-1})||^2]\leq \frac{2(f(\boldsymbol v_{0})-f(\boldsymbol v_{*}))}{ M\tilde B\eta\alpha\sum_{t=1}^{T}\rho_{t-1}} +\\
\nonumber &
\frac{\eta^2 L^2}{\tilde B\sum_{t=1}^{T}\rho_{t-1}} \sum_{t=1}^{T}\rho_{t-1}\bigg[V^2\bigg(\sum_{b=0}^{\tilde B -1}\frac{\mu(\mu^b-1)}{\mu-1}+\tilde B\frac{\mu(\mu^D-1)}{\mu-1}\bigg)+\\\nonumber &M\tilde B^2\sigma^2\sum_{j=t-1-D'}^{t-2}\rho_{j-1}^2\bigg]+\frac{L\eta V^2}{\alpha\sum_{t=1}^{T}\rho_{t-1}} \sum_{t=1}^{T}\rho_{t-1}^2
\end{align}
\text{where $\tilde B=pB$ and $\boldsymbol{v}_*$ is the global optimum of the objective function in Eq.~\ref{intro_1}.} 
\end{theorem}

We denote the expectation of all random variables in Alg.~\ref{alg22} by $\mathop{\mathbb{E}}[.]$. Theorem 1 shows that the weighted average of the $l_2$ norm of all gradients $||\nabla f(\boldsymbol v_{t-1})||^2$ can be bounded, which indicates an ergodic convergence rate. It can be seen that it is possible to achieve speed-up by increasing the number of cores and workers. Nevertheless to reach such speed-up, the learning rates $\eta$ and $\rho_t$ have to be set properly (see Corollary 1).
 
 {\corollary{
By setting the learning rates to be equal and constant:
 \begin{equation}
    \rho^2=\eta^2=\frac{\sqrt{(f(\boldsymbol v_{0})-f(\boldsymbol v_{*}))}}{A\alpha \sqrt{TM\tilde B}}
\end{equation}
such that $A=L V^2\bigg(\frac{1}{\alpha}+\frac{1}{\alpha^2}+ \frac{2L\mu}{(1-\mu)\alpha}\bigg)$, $V>0$  and $\mu$ is a constant where $0<\mu<1$, then the bound in Eq.~\ref{proof_133} and Eq.~\ref{proof_66} can lead to the following bound:

\begin{align}\label{proof_188}
\nonumber T&\geq max\bigg\{\frac{M\tilde B L^2D'^2(f(\boldsymbol v_{0})-f(\boldsymbol v_{*}))}{A^2\alpha^2},\\&\frac{\big(f(\boldsymbol v_{0})-f(\boldsymbol v_{*})\big)\big(\mu(\mu-1)+9L^2\mu(D+1)(\mu^{D+1}-1)\big)^4}{M\tilde BA^2\alpha^2(\mu-1)^8}\bigg\}
\end{align}
and Theorem 1 gives the following  convergence rate:
\begin{equation}\label{proof_200}
 \frac{1}{T}\sum_{t=1}^{T} \mathop{\mathbb{E}}[||\nabla f(\boldsymbol v_{t-1})||^2] \leq 3A\sqrt\frac{f(\boldsymbol v_{0})-f(\boldsymbol v_{*})}{TM\tilde B}
\end{equation}}}

This corollary shows that by setting the learning rates to certain values and setting the number of iterations $T$ to be greater than a bound depending on the maximum delay allowed, a convergence rate of $O(1/\sqrt{TMpB})$ can be achieved and this is delay-independent. The negative effect of using old parameters (asynchronous distribution) and over-writing the shared memory (lock-free parallel) vanish asymptotically. Hence, to achieve speed-up, the number of iterations has to exceed a  bound controlled by the maximum delay parameters, the number of iterations B (line 5 in Alg.~\ref{alg22}), the number of global updates $M$ (line 3 in Alg.~\ref{alg21}) and the number of parallel threads (cores) $p$.
\subsection{Discussion} 

Using Corollary 1, we can derive the result of lock-free parallel optimisation algorithm~\citep{zhao2017lock} and  the asynchronous distributed optimisation algorithm~\citep{lian2015asynchronous} as particular cases. By  setting the number of threads $p=1$ and the number of local update $B=1$, we end up with the distributed asynchronous algorithm presented in~\citep{lian2015asynchronous}. The convergence bound of Corollary 1 then becomes $O(1/\sqrt{TM})$ which is equivalent to that of Corollary 2 in~\citep{lian2015asynchronous}. By synchronising the global learning $D'=0$, setting the master batch size $M=1$ and the number of global iteration $T=1$, we end up with the parallel lock free algorithm presented in~\citep{zhao2017lock}. The convergence bound of Corollary 1 then becomes $O(1/\sqrt{pB})$ which is equivalent to that of Theorem 1 in~\citep{zhao2017lock}. The experiments below will empirically demonstrate these two parallel and distributed particular cases of DPSGD.

Since $D'$ and $D$ are related to the number of workers and cores (threads) respectively, bounding the latter allows speed-up with respect to the number of workers and cores with no loss of accuracy. The satisfaction of Eq.~\ref{proof_188} is guaranteed if:
\begin{equation}
\nonumber T\geq\frac{M\tilde B L^2D'^2(f(\boldsymbol v_{0})-f(\boldsymbol v_{*}))}{A^2\alpha^2}
\end{equation}
and 
\begin{equation}
 \nonumber T\geq\frac{\big(f(\boldsymbol v_{0})-f(\boldsymbol v_{*})\big)\big(\mu(\mu-1)+9L^2\mu(D+1)(\mu^{D+1}-1)\big)^4}{M\tilde BA^2\alpha^2(\mu-1)^8}   
\end{equation}

The first inequality leads to $O(T^{1/2})>D'$. Thus, the upper bound on the number of workers is $O(T^{1/2})$. Since $0<\mu<1$, the second inequality can be written as follows: $O(T^{1/4})\geq \big(\mu(1-\mu)+9L^2\mu(D+1)(1-\mu^{D+1})\big)$. Hence, $O(T^{1/4})\geq D$. Thus, the upper bound on the number of number of cores (threads) is $O(T^{1/4})$. The convergence rate for serial and synchronous parallel stochastic gradient (SG) is consistent with $O(1/\sqrt{T})$~\citep{ghadimi2013stochastic,dekel2012optimal,nemirovski2009robust}. While the workload for each worker running DPSGD is almost the same as the workload of the serial or synchronous parallel SG, the progress done by DPSVG is $Mp$ times faster than that of serial SG. 

In addition to the speed-up, DPSGD allows one to steer the trade-off between multi-core local computation and multi-node communication within the cluster. This can  be done by controlling the parameter $B$. Traditional methods reduce the communication cost by increasing the batch size which decreases the convergence rate, increase local memory load and decrease local input bandwidth. On the contrary, increasing $B$ for DPSGD can increase the speed-up if some assumptions are met (see Theorem 1 and Corollary 1). This ability makes DPSGD easily adaptable to diverse spectrum of large-scale computing systems with no loss of speed-up. 

Denote $Tc$ the communication time need for each master-worker exchange. For simplification, we assume that $Tc$ is fixed and is the same for all nodes. If the time needed for computing one update $Tu\leq Tc$, then the total time needed by the distributed algorithm $DTT=T*(Tu+Tc)$ could be higher than that of the sequential SGD $STT=M*T*Tu$. In such cases, existing distributed algorithms increases the local batch size so that $Tu$ increases, resulting in lower stochastic gradient variance and allowing higher learning rate to be used, hence better convergence rate. This introduces a trade-off between computational efficiency and sample efficiency. Increasing the batch size by a factor of $k$ increases the time need for local computation  by $O(k)$ and reduces the variance proportionally to $1/k$~\citep{bottou2018optimization}. Thus, higher learning rate can be used. However, there is a limit on the size of the learning rate. In another words, maximising the learning speed with respect to the learning rate and the batch size has a global solution. This maximum learning speed can be improved using DPSGD, performing $B$ times less communication steps. For the mini-batch SGD with minibatch
size $G$, the convergence rate can be written as $O(1/\sqrt{GT})$. Since the total number of examples examined is $GT$ and there is only $\sqrt{G}$ times improvement, the convergence speed degrades as mini-batch size increases. The convergence rate of DPSGD with mini-batch $G$ can be easily deduced from Theorem 1 as $O(1/\sqrt{BMGT})$. Hence, $\sqrt{BM}$ better convergence rate than mini-batch SGD and $\sqrt{BM}$ better convergence rate than standard asynchronous SGD with $B$ times less communication. These improvements are studied in the following.

\section{Experiments} \label{sec4}

In this section, we empirically verify the potential speed-up gains expected from the theoretical analysis. First, we apply distributed parallel stochastic variational inference (DPSVI) algorithm on a Latent Dirichlet Allocation (LDA) analysis problem. DPSVI is derived from DPSGD by replacing the SG of SVI by DPSG to scale up the inference computation over a multi-core cluster (see appendix for more details). For the  Latent Dirichlet Allocation analysis problem, we use the SVI algorithm~\citep{hoffman2013stochastic} as benchmark. The evaluation is done on $300,000$ news articles from the \textit{New York Times} corpus. 

Furthermore, we  use DPSGD to scale up the training of DRL algorithm, namely \textit{Advantage Actor Critic (A2C)} algorithm, implementing \textit{highly scalable A2C (HSA2C)} (details in the appendix). We compare HSA2C against other distributed A2C implementations using a testbed of six Atari games and demonstrate an average training time of 21.95 minutes compared to over 13.75 hours by the baseline A3C. In particular, HSA2C shows a significant speed-up on Space invaders with learning time below 10 minutes compared to the 30 minutes achieved by the best competitor.

\subsection{Variational inference} 
The development of the proposed DPSVI algorithm follow from  DPSGD, but in the context of VI. In Appendix ~\ref{applica}, we characterise the entire family of models where DPSVI is applicable, which is shown to be equivalent to the models for which SVI applies. Next, DPSVI is derived from DPSGD. Finally, we derive an asynchronous distributed lock-free parallel inference algorithm for LDA as a case study for DPSVI. 

\textbf{Datasets:} We use the  \textit{NYTimes} corpus ~\citep{Lichman:2013} containing $300,000$ news articles from the \textit{New York Times} corpus. The data is pre-processed by removing all the words not found in a dictionary containing $102,660$ most frequent words - see~\citep{Lichman:2013} for more information. We reserve $5,000$ documents from \textit{NYTimes} data as a validation set and another $5,000$ documents as a testing set.

\textbf{Performance:} The performance of the LDA model is assessed using a model fit measure, {\it perplexity}, which is defined as the geometric mean of the inverse marginal probability of each word in the held-out set of documents~\citep{blei2003latent}. We also compute the running time speed-up (TSP) ~\citep{lian2015asynchronous} defined as 
\begin{align}\label{df}
TSP&=\frac{ T( \text{SVI}) }{ T( \text{DPSVI}) }
\end{align}
where $T(\cdot)$ denotes the running time and is taken when both models achieve the same final held-out perplexity of 5000 documents.

\textbf{Parameters:}  In all experiments, the LDA number of topics is $K = 50$. SVI LDA  is run on the training set for $\kappa\in \{0.5, 0.7, 0.9\}$, $\tau_0\in\{1, 24, 256, 1024\}$, and $batch\in \{16, 64, 256, 1024\}$. The best performing parameters $batch=1024$, $\kappa=0.5$ and $\tau_0=1$ providing preplexity of $5501$ are used (Table~1 in~\citep{smohamad} summarises the best settings with the resulting perplexity on the test set). As for the DPSGD LDA version, the local learning rate $G$ (see Eq.~\ref{equ14}) is set to $64$ and $M$ equal to $16$. We evaluate a range of learning rates $\eta=\rho\in\{0.2,0.1,0.05, 0.01\}$ where $M$, $p$ and $B$ are set to $1$. The best learning rate $0.1$ providing held-out perplexity of $5501$ was used. For different $B$, $M$ and $p$, the learning rate is changed according to Corollary 1:
\begin{equation}\label{exp1}
\rho'=\rho\bigg(\frac{pBM}{p'B'M'}\bigg)^{0.25} =\frac{0.1}{(p'B'M')^{0.25}}
\end{equation}
 All DPSGD LDA experiments were performed on a high-performance computing (HPC) environment using  message passing interface (MPI) for python (MPI4py). The \textit{cluster} consists of 10 nodes, including the head node, with each node being a 1-sockets-6-cores-2-thread processor.
\subsubsection{Node speed-up}

\begin{figure}[tb]
\centering
\includegraphics[height=5.5cm,width=0.6\textwidth]{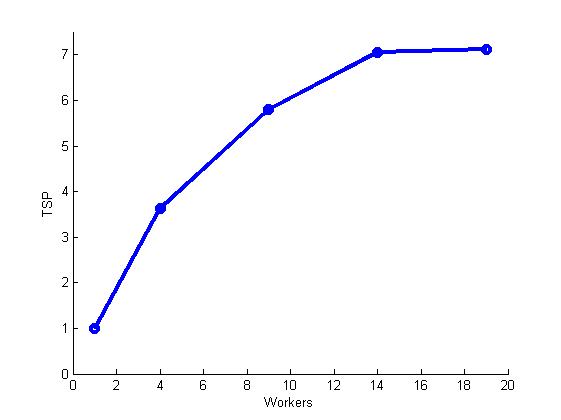}
\caption{LDA analysis: Running time speed-up (TSP) with respect to the number of workers.}
\label{fig2}
\end{figure}

Here, we study the speed-up of DPSVI with respect to the number of workers where $p=1$ and $B=1$. DSPSVI LDA is then compared against serial SVI ($B=1$, $p=1$ and $nW=1$). We run DPSVI for various numbers of workers $nW\in \{4, 9, 14, 19 \}$. The number of nodes is  $nW$ as long as $nW$ is less than $9$. As $nW$ becomes higher than the available nodes, the processors' cores of nodes are employed as workers until all cores (threads) of each node are used i.e., $9\times 12=108$. The batch size $M$ is fixed to $36$. Figure~\ref{fig2} summarises the total speed-up (i.e., TSP measured at the end  of the algorithm) with respect to the number of workers where the achieved pre-perplexity is almost the same. The result shows linear speed-up as long as the number of workers is less than $14$. Then, linear speed-up slowly converts to sub-linear and is expected to drop for higher number of workers due to reaching the maximum communication bandwidth.  
\subsubsection{Thread speed-up} 
\begin{figure}[tb]
\centering
\includegraphics[height=5.5cm,width=0.6\textwidth]{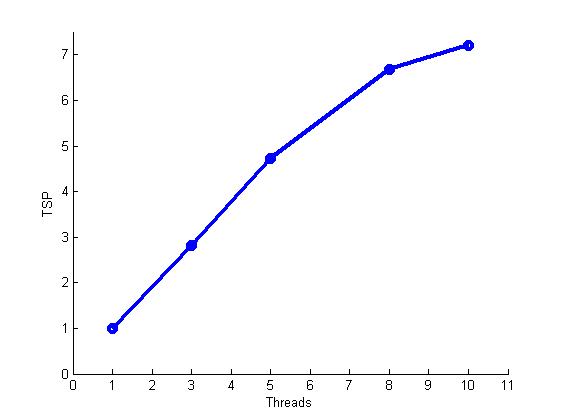}
\caption{LDA analysis: Running time speed-up (TSP) with respect to the number of threads.}
\label{fig3}
\end{figure}
In this section, we study the speed-up of DPSVI with respect to the number of threads where $nW=1$. We empirically set $B$ to $15$. Similar to the node-related speed-up analysis, experiments are run for different $p\in\{3,5,8,10\}$. Then, DPSVI is compared against serial SVI. The results are shown in Fig.~\ref{fig3}. The outcome shows linear speed-up as long as the number of threads is less than $8$. Then, the speed-up slowly converts to sub-linear and is expected to become  worse for higher number of threads. This drop in the speed-up is due to hardware communication and other factors affecting the CPU power.  

\subsubsection{Node-thread speed-up}

\begin{figure}[tb]
\centering
\includegraphics[height=6.5cm,width=0.8\textwidth]{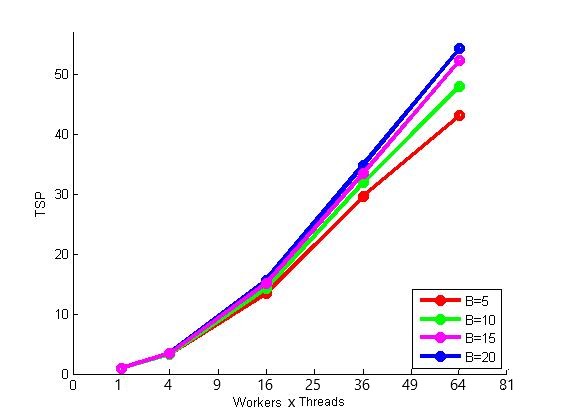}
\caption{LDA analysis: Running time speed-up (TSP) with respect to the number of workers and threads.}
\label{fig4}
\end{figure}

\begin{figure}[tb]
\centering
\includegraphics[height=6.5cm,width=0.8\textwidth]{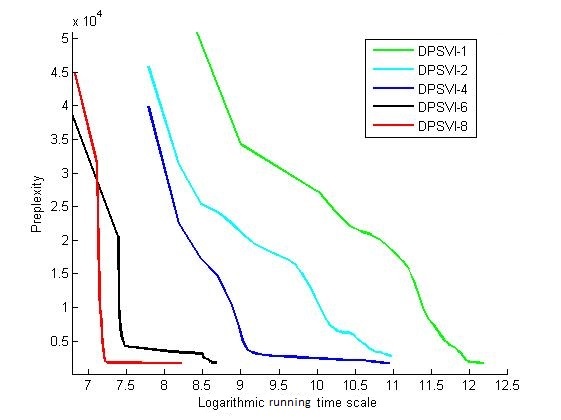}
\caption{LDA analysis using DPSVI: perplexity (model fit) with respect to running logarithmic time in seconds.}
\label{fig5}
\end{figure}

Finally, we study the speed-up of DPSVI with respect to the number of nodes and threads. To simplify the experiments, we take the number of cores to be equal to the number of nodes. Experiments are run for different $p=nW\in\{2,4,6,8\}$. We also present results with different $B\in\{5,10,15,20\}$ in order to show the effect of steering the trade-off between local computation and communication. DPSVI is compared against serial SVI and the results are shown in Fig.~\ref{fig4}. The result shows speed-up whose speed slows down as the number of threads and nodes exceed 6. This is due to communication and other hardware factors. However, the rate of this slowing down for higher $B$ is less significant which illustrates the advantage of reducing the communication overhead when reaching its ceiling point. Note that for very high number of workers, increasing $B$ might not be very helpful as our theoretical results show that high $B$ tightens the bound on the number of workers allowed for the speed-up to holds. Fig.~\ref{fig5} reports the perplexity on the training set with respect to running time in seconds (logarithmic scale) with $B=15$. Five curves are drawn for different nodes-threads number, where DPSVI-n denotes our DPSVI with $n$ nodes and threads. The convergence and speed-up of DPSVI are clearly illustrated. 

\subsection{Deep reinforcement learning}

We use six different Atari games to study the performance gains that can be achieved by the proposed HSA2C algorithm using the Atari 21600 emulator~\citep{bellemare2013arcade} provided by the OpenAI Gym framework~\citep{brockman2016openai}. This emulator is one of the most commonly used benchmark
environments for RL algorithms. Here, we use Pong, Boxing, Seaquest Space invaders, Amidar and Qbert which have been included in related work~\citep{mnih2016asynchronous,adamski2017atari,adamski2018distributed,babaeizadeh2016reinforcement}. These games are used to evaluate the effects of  reducing the communication bottleneck when using an increasingly higher number of steps, $B$, with different numbers of nodes. We also study the speed-up achieved by HSA2C with respect to the number of nodes. Finally, we compare the performance reported by various state-of-the-art algorithms ~\citep{mnih2016asynchronous,adamski2017atari,adamski2018distributed,babaeizadeh2016reinforcement}. 

\subsubsection{Implementation details} 
HSA2C has been implemented and tested on a high-performance computing (HPC) environment using message passing interface (MPI) for Python (MPI4py 3.0.0) and Pytorch 0.4.0. Our cluster consists of $60$ nodes consisting of $28$ 2.4 GHz CPUs per node. In our experiments, we used the same input pre-processing as~\citep{mnih2015human}. Each experiment was repeated $5$ times (each with an action repeat of 4) and the average results are reported. The agents used the neural network architectures described in ~\citep{mnih2013playing}: a convolutional layer with 16 filters of size 8 x 8 with stride 4, followed by a convolutional layer with with 32 filters of size 4 x 4 with stride 2, followed by a fully connected layer with 256 hidden units. All three hidden layers were followed by a rectifier nonlinearity. The network has two sets of outputs – a softmax output with one entry per action representing the probability of selecting the action and a single linear output representing the value function. Local learning rate, mini-batch size and the optimiser setting are contrasted with those reported in~\citep{adamski2018distributed} in order to provide a fair comparison with their asynchronous mini-batch implementation. The global learning rate was set to $0.01$ for the SGD optimiser with $0.5$ momentum. The global batch was set to the number of utilised nodes.

\subsubsection{Speed-up analysis}\label{anasp} 

\begin{figure}[tb]
\centering
\includegraphics[width=\textwidth]{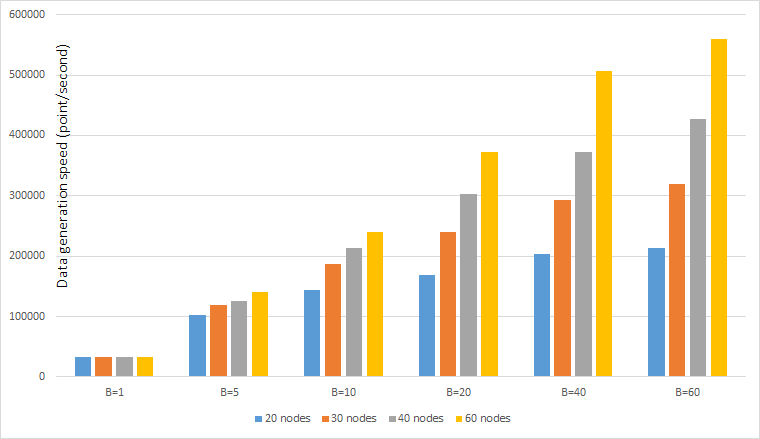}
\caption{The average data generation speed of HSA2C measured in points per second within 30 minutes run on Space invader game.}
\label{fig1i}
\end{figure}

\begin{figure*}[ht]
\subfigure[Pong (reference solution:$-3$)]{\includegraphics[width = 0.5\textwidth, height=0.38\textwidth]{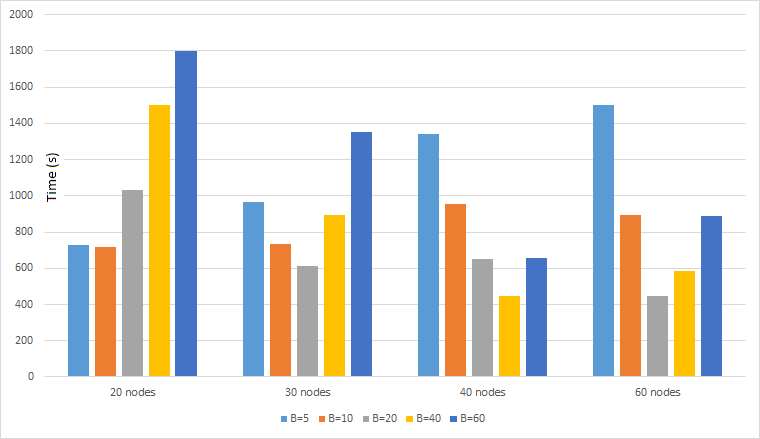}\label{fig11}}
\subfigure[Boxing (reference solution: $72.8$)]{\includegraphics[width = 0.5\textwidth, height=0.38\textwidth]{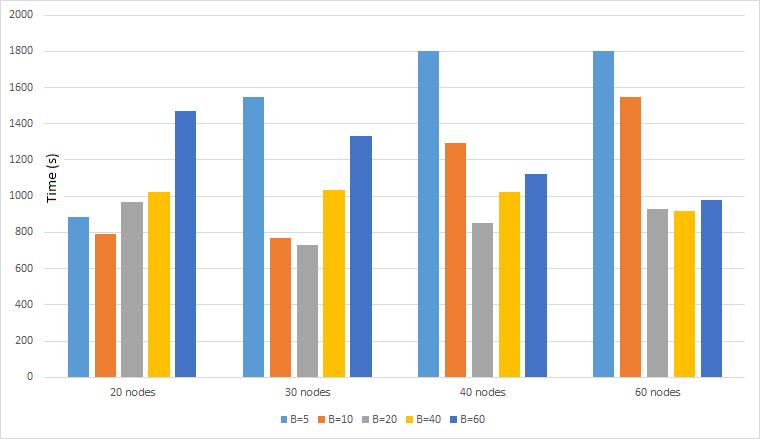}\label{fig12}}\\
\subfigure[Seaquest (reference solution: $1062$)]{\includegraphics[width = 0.5\textwidth, height=0.38\textwidth]{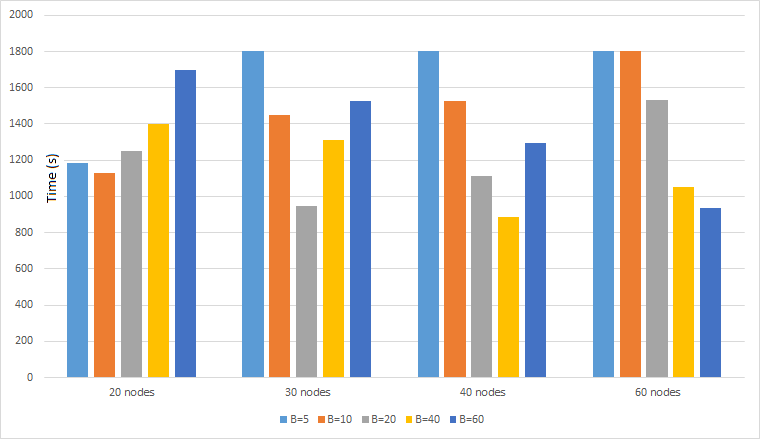}\label{fig13}}
\subfigure[Space invader (reference solution: $664$)]{\includegraphics[width = 0.5\textwidth, height=0.38\textwidth]{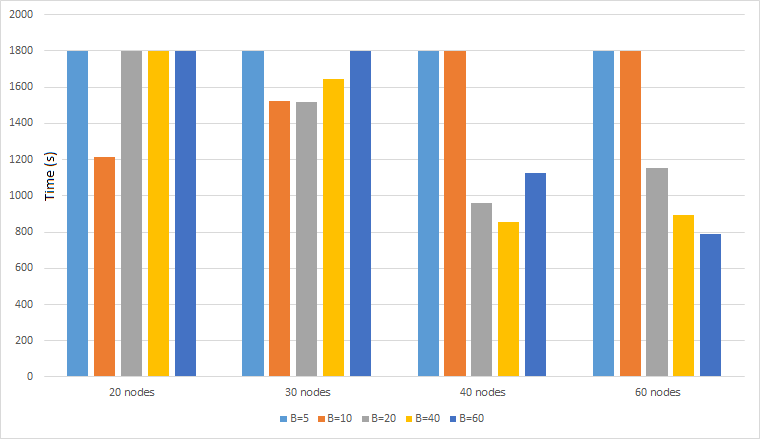}\label{fig14}}\\
\subfigure[Amidar (reference solution: $186$)]{\includegraphics[width = 0.5\textwidth, height=0.38\textwidth]{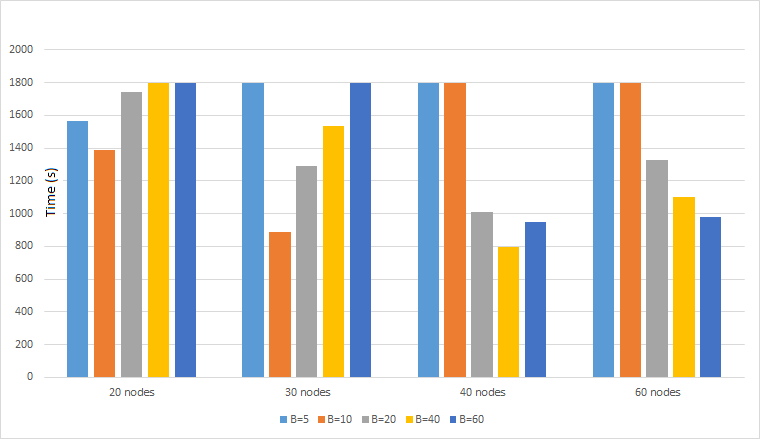}\label{fig15}}
\subfigure[Qbert (reference solution: $875$)]{\includegraphics[width = 0.5\textwidth, height=0.38\textwidth]{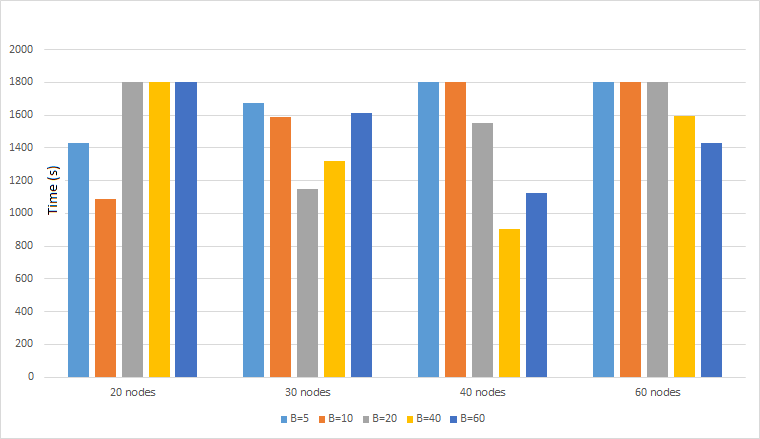}\label{fig16}}
\caption{The time (in seconds) required to reach reference solution (the highest score with $B=1$ in a 30 minutes run) over a range of node numbers and $B$.}\label{fnew}
\end{figure*}

In this section, we study the effect of $B$ on various aspects of the scalability of HSA2C with respect to the number of nodes. Figure~\ref{fig1i} shows the average speed of data generated by the distributed actors measured in data-points per seconds on the Space invader game. Comparable figures can be obtained for other games, but they are not reported here. It is noticeable that at $B=1$, the speed of data generation is about the same as the number of nodes increases. This is due to the communication cost which increases with the number of nodes used. That is, the expected waiting time of each node's exchange with the master increases. Increasing $B$ will reduce the number of exchanges while performing the same number of updates locally. This is illustrated in Fig.~\ref{fig1i} where the data generation speed increases with the number of nodes as $B$ increases. 

Figure~\ref{fnew} shows the time (in seconds) required to reach the highest score of Pong, Boxing, Seaquest, Space invader, Amidar and Qbert reached with $B= 1$ in a 30-minute run over different number of nodes and $B$. The aim of these figures is to demonstrate the potential performance gains achieved by HSA2C as $B$ increases in comparison to distributed deep reinforcement learning (DDRL) A3C, which is algorithmically equivalent to HSA2C when $B=1$ (our baseline). In order to produce these figures, we initially carried out a search to empirically determine the highest score that can be achieved by HSA2C within a time period of $30$ minutes when $B=1$. The search for the highest score is performed on four different cluster sizes: $20$, $30$, $40$, and $60$. The figures present the time required for various HSA2C parameters to reach that benchmark score. These experimental results clearly show the impact of $B$ on the communication costs and confirm the findings of Fig.~\ref{fig1i}. By using a larger number of nodes, more communication exchanges are required for each update and performing more local update (i.e., increasing $B$) reduces the communication exchanges needed to reach certain score without much sacrificing the learning performance. Thus, a better speed-up can be achieved. On the other hand, with a smaller number of nodes, increasing $B$ does not make a significant difference in reducing the communication whilst the negative effect of an increased variance becomes significant as the size of the learning batch becomes smaller (depending on the number of nodes). Overall, there is evidence for a variance-communication trade-off controlled by $B$. 

The multiple local updates $B$ mitigates the speed-up limit caused by higher communication cost with higher number of nodes. Choosing the right $B$ for different numbers of nodes allows HSA2C to scale better than HSA2C when $B=1$. For all the six games, we have found that increasing the number of nodes over $40$ does not lead to better performance. This is due to the higher variance entailed by a higher $B$. That is, the performance improvement coming from the communication reduction is overtaken by the entailed variance when using more than $40$ nodes. This limit could be overcome in two different ways, either by decreasing the communication without further increasing the variance or by directly mitigating the variance problems. 

\subsubsection{Comparison}

We show the effectiveness of the proposed approach by comparing it against similar scalable Actor-critic optimisation approaches. The most similar work in speeding up Atari games training is presented in~\citep{adamski2018distributed,mnih2016asynchronous,adamski2017atari,babaeizadeh2016reinforcement}. The algorithm in~\citep{adamski2018distributed} (DDRL A3C) is a particular case of HSA2C, where the communication is synchronised and the number of iterations of per-worker loop is set to one ($B=1$). GA3C is a hybrid GPU/CPU algorithm  which is a flavour of A3C focusing on batching the data points in order to better utilise the massively parallel nature of GPU computations. This is similar to the single-node algorithm called BA3C~\citep{adamski2017atari}.

Table~\ref{tab001} presents the best score and time (in minutes) HSA2C archives using the best $B$ values found in Fig.~\ref{fnew} in comparison to the competitors. The reported scores are taken from the original papers. As GA3C, BA3C and A3C are parallel single-node algorithms, their experimental settings are not comparable to ours. This comparison shows that our approach achieves a better score than all competitors. In particular, we achieve an average score of $665$ in $30.63$ minutes average time using $560$ total CPU cores compared to the DDRL A3C score of $650$ in $82.5$ average time with $778$ total CPU cores. Most importantly, this comparison validates the effectiveness of our proposed approach to reduce communication while preserving performance. This is clearly shown in the comparison between our approach and our implementation of DDRL A3C (the top competitor in Tab.~\ref{tab001}) using the same setting (see Sec.~\ref{anasp}). 
 
For this study, we have decided not to include GPU-based implementations such as~\citep{stooke2018accelerated} as our focus here is on CPU-enabled methods. However, HSA2C is generic and lends itself to GPU-based implementations whereby each node consists of multiple CPUs and a GPU. In such a case, local computation and simulation can be done using CPUs/GPU units, where our multiple local update approach can further speed up the standard DA3C communication-based~\citep{stooke2018accelerated}. The empirical work reported here provides an initial validation of the underlying idea. 

\begin{table}[!t]
\caption{Best scores and the corresponding time in minutes achieved by HSA2C using the best $B$ from Table~\ref{tab001} and  considering $20$, $30$, $40$ and $60$ nodes compared to the best reported results by competitors.}
\label{tab001}
  \begin{adjustbox}{width=1\textwidth,center}
\begin{tabular}{|l|l|l|l|l|l|l|l|l|l|l|}\hline
Algorithms& Pong&Boxing&Seaquest&Space invaders  & Amidar& Qbert \\\hline
HSA2C&$20~(63m)$&$98~(19.8m)$&$1858~(21.7m)$  & $684~(17.9m)$&$251~(26m)$  &$1210~(20m)$\\(20 nodes)&&&&&&\\\hline

HSA2C&$19~(40m)$&$98~(19.8m)$&$1874~(19.1m)$&$671~(14.7m)$  & $234~(15m)$& $1420~(27m)$ \\(30 nodes)&&&&&& \\\hline

HSA2C&$20~(32.6m)$&$95~(20.7m)$&$1894~(20.5m)$&$681~(9.9m)$ & $282~(20m)$& $1620~(28m)$ \\(40 nodes)&&&&&&\\\hline

HSA2C & $19~(38m)$&$96~(23.2m)$&$1874~(22.3m)$&$667~(9.4m)$  & $243~(24m)$& $800~(30m)$  \\(60 nodes)&&&&&&\\\hline
DDRL A3C &$20~(240m)$&$98~(30m)$&$1832~(30m)$&$650~(30m)$  & ----& ---- \\\hline
GA3C &$18~(60m)$&$92~(120m)$ &$1706~(1440m)$&$600~(1440m)$  & $218~(1440m)$& $395~(30m)$ \\\hline
BA3C & $17~(1440m)$& ----& $1840~(1440m)$ &$700~(1440m)$  & ----& ----  \\\hline
A3C&$20~(480m)$ &$95~(660m)$&$2300~(1440m)$&$1400~(900m)$   & $280~(1440m)$& $400~(30m)$   \\\hline
\end{tabular}
\end{adjustbox}
\end{table}

\section{Conclusion}\label{sec7}

We have proposed a novel asynchronous distributed and lock-free parallel optimisation algorithm. The algorithm has been implemented on a computer cluster with multi-core nodes. Both theoretical and empirical results have shown that DPSGD  leads to speed-up on non-convex problems. The paper shows how DPSVI and HSA2C have been derived from DPSGD. Both are an asynchronous distributed and lock-free parallel implementation for respectively stochastic variational inference (SVI) and advantage actor critic (A2C). Empirical results have allowed to validate the theoretical findings and to compare against similar state-of-the-art methods.

Going forward, further improvements and validations could be achieved by pursuing research along five directions: (1) employing variance reduction techniques to improve the convergence rate (from sub-linear to linear) while guaranteeing multi-node and multi-core speed-up; (2) proposing a framework enabling dynamic trade-offs between local computation and communication; (3) proposing techniques to improve the local optimum of the distributed parallel algorithms; (4) applying DPSVI to other members of the family of models stated in the appendix; (5) applying DPSGD to other large-scale deep learning problems. 

\subsubsection*{Acknowledgement}

The authors thank the HPC team at Warwick University for providing the computational resources to run the experiments.

\section*{Declarations}

Some journals require declarations to be submitted in a standardised format. Please check the Instructions for Authors of the journal to which you are submitting to see if you need to complete this section. If yes, your manuscript must contain the following sections under the heading `Declarations':

\begin{itemize}
\item Funding: This work was partly funded through the European Horizon 2020 Framework Programme under grant 687691 related to the Project: PROTEUS: Scalable Online Machine Learning for Predictive Analytics and Real-Time Interactive Visualization. 

\item Conflict of interest/Competing interests: None.
\item Ethics approval: Not applicable. 
\item Consent to participate: Not applicable.
\item Consent for publication: Not applicable.
\item Availability of data and materials: Not applicable.
\item Code availability: Code could be make available at a later stage. 
\item Authors' contributions: S. Mohammad developed and implemented the concepts. He wrote the draft and supported the subsequent revision. H. Bouchachia initiated the idea of the research and followed up its development. He worked on the manuscript throughout the publication process. H. Alamri contributed to the implementation of the reinforcement learning part of the manuscript.
\end{itemize}

\bibliography{refer.bib}
\newpage

\onecolumn

\section{Proofs}

Let $q(\boldsymbol x)=\frac{1}{n}\sum_{i=1}^{n}||\nabla f_i(\boldsymbol x )||^2$. We have $\mathbb{E}_i[||\nabla f_i(\boldsymbol x )||^2]=q(\boldsymbol x )$. Hence, $\mathbb{E}_i[q(\boldsymbol x )]=\mathbb{E}_i[||\nabla f_i(\boldsymbol x )||^2]$. Taking the full expectation on both sides, we get $\mathbb{E}[q(\boldsymbol x)]=\mathbb{E}[||\nabla f_i(\boldsymbol x )||^2]$. It can be proven that:
\begin{lemma}
\begin{equation}\label{proof_55}
\mathbb{E}_t[q(\boldsymbol{ \hat u}_{t,m,j})]<\mu \mathbb{E}_t[q(\boldsymbol{ \hat u}_{t,m,j+1})]
\end{equation}
given $\mu$ and $\eta$ satisfying 
\begin{equation}\label{proof_6}
\frac{1}{1-\eta-\frac{9\eta(D+1)L^2(\mu^{D+1}-1)}{\mu-1}}\leq \mu 
\end{equation}
where $\mathbb{E}_t[.]$ denotes $\mathbb{E}_{i_{t,*,*},S_{t,*,*}}[.]$. The proof can be derived from that in~\citep{zhao2017lock} using Assumptions (1) and (5). The stars means for all, $\forall$.
\end{lemma}

\textbf{Proof to Theorem 1}
From the smoothness Assumption (1), we have:
\begin{align}\label{proof_1}
\nonumber f(\boldsymbol v_{t})-f(\boldsymbol v_{t-1})&\leq  \langle\nabla f(\boldsymbol v_{t-1}),\boldsymbol v_t-\boldsymbol v_{t-1}\rangle+\frac{L}{2}||\boldsymbol v_t-\boldsymbol v_{t-1}||^2\\\nonumber&=-\langle\nabla f(\boldsymbol v_{t-1}),\eta\rho_{t-1}\sum_{m=1}^{M}\sum_{b=0}^{\tilde B-1}S_{t,m,b}\nabla f_{i_{t,m,b}}(\boldsymbol{ \hat u}_{t-\tau_{t,m},m,b})\rangle\\&+\frac{L}{2}||\eta\rho_{t-1}\sum_{m=1}^{M}\sum_{b=0}^{\tilde B-1}S_{t,m,b}\nabla f_{i_{t,m,b}}(\boldsymbol{ \hat u}_{t-\tau_{t,m},m,b})||^2
\end{align}
where the last equality uses the update in Eq.~\eqref{equ_2_1}. Taking  expectation of the above inequality with respect to  $i_{t,*,*}$, and $S_{t,*,*}$, we obtain: 
\begin{align}\label{proof_2}
\nonumber E_t[f(\boldsymbol v_{t})]-f(\boldsymbol v_{t-1})&\leq-M\tilde B\eta\rho_{t-1}\langle\nabla f(\boldsymbol v_{t-1}),\frac{1}{M\tilde B}\sum_{m=1}^{M}\sum_{b=0}^{\tilde B-1}S\nabla f(\boldsymbol{ \hat u}_{t-\tau_{t,m},m,b})\rangle\\&+\frac{L\eta^2\rho_{t-1}^2}{2}E_t||\sum_{m=1}^{M}\sum_{b=0}^{\tilde B-1}S_{t,m,b}\nabla f_{i_{t,m,b}}(\boldsymbol{ \hat u}_{t-\tau_{t,m},m,b})||^2
\end{align}
 where we used Assumptions (2), (3), (4) and (6). Since $S$ is a strictly definite matrix with the largest \textit{eigenvalue} less or equal than $1$ and the minimum \textit{eigenvalue} is $\alpha>0$ and from the fact$\langle a,b \rangle=\frac{1}{2}(||a||^2+||b||^2-||a-b||^2)$, we have: 
 \begin{align}\label{proof_3}
\nonumber E_t[f(\boldsymbol v_{t})]-f(\boldsymbol v_{t-1})&\leq-\frac{M\tilde B\eta\rho_{t-1}\alpha}{2}\bigg(||\nabla f(\boldsymbol v_{t-1})||^2+||\frac{1}{M\tilde B}\sum_{m=1}^{M}\sum_{b=0}^{\tilde B-1}\nabla f(\boldsymbol{ \hat u}_{t-\tau_{t,m},m,b})||^2-\\\nonumber&\underbrace{||\nabla f(\boldsymbol v_{t-1})-\frac{1}{M\tilde B}\sum_{m=1}^{M}\sum_{b=0}^{\tilde B-1}\nabla f(\boldsymbol{ \hat u}_{t-\tau_{t,m},m,b})||^2}_{\text{H1}}\bigg)+\\&\frac{L\eta^2\rho_{t-1}^2}{2}\underbrace{E_t||\sum_{m=1}^{M}\sum_{b=0}^{\tilde B-1}S_{t,m,b}\nabla f_{i_{t,m,b}}(\boldsymbol{ \hat u}_{t-\tau_{t,m},m,b})||^2}_{\text{H2}}
\end{align}

Next, we obtain an upper bound for  H1. Using the triangular inequality and Assumption (1), we can write the following:
\begin{align}\label{proof_4}
\nonumber H1&=||\nabla f(\boldsymbol v_{t-1})-\frac{1}{M\tilde B}\sum_{m=1}^{M}\sum_{b=0}^{\tilde B-1}\nabla f(\boldsymbol{ \hat u}_{t-\tau_{t,m},m,b})||^2\\\nonumber&\leq\frac{1}{M\tilde B}\sum_{m=1}^M\sum_{b=0}^{\tilde B-1}||\nabla f(\boldsymbol v_{t-1})-\nabla f(\boldsymbol{ \hat u}_{t-\tau_{t,m},m,b})||^2\\&\leq\frac{L^2}{\tilde B}\sum_{b=0}^{\tilde B-1}||\boldsymbol v_{t-1}-\boldsymbol{ \hat u}_{t-\tau_{t,y},y,b}||^2
\end{align}
where $y=argmax_{m\in\{1,...M\}}||\boldsymbol v_{t-1}-\boldsymbol{ \hat u}_{t-\tau_{t,m},m,b}||^2$. Using triangular inequality and the updates in Eq.~\eqref{equ_2_1}, we obtain:
\begin{align}\label{proof_5}
\nonumber H1&\leq\frac{L^2}{\tilde B}\sum_{b=0}^{\tilde B-1}\bigg(||\boldsymbol v_{t-1}-\boldsymbol{u}_{t-\tau_{t,y},y,a(b)}||^2+\eta^2||\sum_{j=a(b)}^{b-1}\nabla f_{i_{t,y,j}}(\boldsymbol{\hat u}_{t-\tau_{t,y},y,j})||^2\bigg)\\\nonumber&\leq \frac{L^2}{\tilde B}\sum_{b=0}^{\tilde B-1}\bigg(||\boldsymbol v_{t-1}-\boldsymbol{v}_{t-1-\tau_{t,y}}||^2+\eta^2||\sum_{j=0}^{b-1}\nabla f_{i_{t,y,j}}(\boldsymbol{\hat u}_{t-\tau_{t,y},y,j})||^2+\eta^2||\sum_{j=a(b)}^{b-1}\nabla f_{i_{t,y,j}}(\boldsymbol{\hat u}_{t-\tau_{t,y},y,j})||^2\bigg)\\\nonumber&= \frac{L^2}{\tilde B}\sum_{b=0}^{\tilde B-1}\bigg(\eta^2\underbrace{||\sum_{j=t-1-\tau_{t,y}}^{t-2}\rho_{j-1}^2\sum_{m=1}^{M}\sum_{b=0}^{\tilde B-1}\nabla f_{i_{j,m,b}}(\boldsymbol{\hat u}_{j-\tau_{j,m},m,b})||^2}_{\text{H11}}+\eta^2||\sum_{j=0}^{b-1}\nabla f_{i_{t,y,j}}(\boldsymbol{\hat u}_{t-\tau_{t,y},y,j})||^2\\&+\eta^2||\sum_{j=a(b)}^{b-1}\nabla f_{i_{t,y,j}}(\boldsymbol{\hat u}_{t-\tau_{t,y},y,j})||^2\bigg)
\end{align}

Using the triangular inequality again, we have the following:
\begin{align}\label{proof_6}
\nonumber H11&=||\sum_{j=t-1-\tau_{t,y}}^{t-2}\rho_{j-1}^2\sum_{m=1}^{M}\sum_{b=0}^{\tilde B-1}\nabla f_{i_{j,m,b}}(\boldsymbol{\hat u}_{j-\tau_{j,m},m,b})||^2\\\nonumber &=||\sum_{j=t-1-\tau_{t,y}}^{t-2}\rho_{j-1}^2\sum_{m=1}^{M}\sum_{b=0}^{\tilde B-1}\bigg(\nabla f_{i_{j,m,b}}(\boldsymbol{\hat u}_{j-\tau_{j,m},m,b})-\nabla f(\boldsymbol{\hat u}_{j-\tau_{j,m},m,b})+\nabla f(\boldsymbol{\hat u}_{j-\tau_{j,m},m,b})\bigg)||^2\\\nonumber&\leq||\sum_{j=t-1-\tau_{t,y}}^{t-2}\rho_{j-1}^2\sum_{m=1}^{M}\sum_{b=0}^{\tilde B-1}\bigg(\nabla f_{i_{j,m,b}}(\boldsymbol{\hat u}_{j-\tau_{j,m},m,b})-\nabla f(\boldsymbol{\hat u}_{j-\tau_{j,m},m,b})\bigg)||^2\\ &+||\sum_{j=t-1-\tau_{t,y}}^{t-2}\rho_{j-1}^2\sum_{m=1}^{M}\sum_{b=0}^{\tilde B-1}\nabla f(\boldsymbol{\hat u}_{j-\tau_{j,m},m,b})||^2
\end{align}
By taking the expectation on both sides with respect to all random variables associated with $k\in\{t-1-\tau_{t,y},...,t-2\}$  and using Assumptions (6), (5) and (4), we obtain:
\begin{align}\label{proof_7}
\nonumber E_k[H11]&\leq\sum_{j=t-1-\tau_{t,y}}^{t-2}\rho_{j-1}^2E_k||\sum_{m=1}^{M}\sum_{b=0}^{\tilde B-1}\bigg(\nabla f_{i_{j,m,b}}(\boldsymbol{\hat u}_{j-\tau_{j,m},m,b})-\nabla f(\boldsymbol{\hat u}_{j-\tau_{j,m},m,b})\bigg)||^2\\\nonumber &+D'\sum_{j=t-1-\tau_{t,y}}^{t-2}\rho_{j-1}^2E_k||\sum_{m=1}^{M}\sum_{b=0}^{\tilde B-1}\nabla f(\boldsymbol{\hat u}_{j-\tau_{j,m},m,b})||^2\\&\leq M\tilde B\sigma^2\sum_{j=t-1-\tau_{t,y}}^{t-2}\rho_{j-1}^2+D'\sum_{j=t-1-\tau_{t,y}}^{t-2}\rho_{j-1}^2E_k||\sum_{m=1}^{M}\sum_{b=0}^{\tilde B-1}\nabla f(\boldsymbol{\hat u}_{j-\tau_{j,m},m,b})||^2
\end{align}
Taking the full expectation, we have
\begin{align}\label{proof_8}
\nonumber E[H1]&\leq \frac{L^2\eta^2}{\tilde B}\sum_{b=0}^{\tilde B-1}\bigg(M\tilde B\sigma^2\sum_{j=t-1-\tau_{t,y}}^{t-2}\rho_{j-1}^2+D'\sum_{j=t-1-\tau_{t,y}}^{t-2}\rho_{j-1}^2E_k||\sum_{m=1}^{M}\sum_{b=0}^{\tilde B-1}\nabla f(\boldsymbol{\hat u}_{j-\tau_{j,m},m,b})||^2\\&+||\sum_{j=0}^{b-1}\nabla f_{i_{t,y,j}}(\boldsymbol{\hat u}_{t-\tau_{t,y},y,j})||^2+||\sum_{j=a(b)}^{b-1}\nabla f_{i_{t,y,j}}(\boldsymbol{\hat u}_{t-\tau_{t,y},y,j})||^2\bigg)
\end{align}
Using Assumption (4) and Lemma 1 we have the following:
\begin{align}\label{proof_9}
\nonumber E[H1]&\leq \frac{L^2\eta^2}{\tilde B}\sum_{b=0}^{\tilde B-1}\bigg(M\tilde B\sigma^2\sum_{j=t-1-\tau_{t,y}}^{t-2}\rho_{j-1}^2+D'\sum_{j=t-1-\tau_{t,y}}^{t-2}\rho_{j-1}^2E||\sum_{m=1}^{M}\sum_{b=0}^{\tilde B-1}\nabla f(\boldsymbol{\hat u}_{j-\tau_{j,m},m,b})||^2\\&+\frac{\mu(\mu^b-1)}{\mu-1}V^2+\frac{\mu(\mu^D-1)}{\mu-1}V^2\bigg)
\end{align}

Using Assumption (4), the full expectation of $H2$ can be bounded as follows:
\begin{equation}\label{proof_10}
E[H2]\leq\frac{L\eta^2\rho_{t-1}^2}{2}M\tilde B V^2
\end{equation}
By taking the full expectation of Eq.~\ref{proof_3} and applying the upper bound of $E[H1]$ and $E[H2]$, we obtain:
\begin{align}\label{proof_11}
\nonumber E[f(\boldsymbol v_{t})]-E[f(\boldsymbol v_{t-1})]&\leq-\frac{M\tilde B\eta\rho_{t-1}\alpha}{2}\bigg[E[||\nabla f(\boldsymbol v_{t-1})||^2]+E[||\frac{1}{M\tilde B}\sum_{m=1}^{M}\sum_{b=0}^{\tilde B-1}\nabla f(\boldsymbol{ \hat u}_{t-\tau_{t,m},m,b})||^2]-\\\nonumber &\frac{L^2\eta^2}{\tilde B}\sum_{b=0}^{\tilde B-1}\bigg(M\tilde B\sigma^2\sum_{j=t-1-\tau_{t,y}}^{t-2}\rho_{j-1}^2+D'\sum_{j=t-1-\tau_{t,y}}^{t-2}\rho_{j-1}^2E||\sum_{m=1}^{M}\sum_{b=0}^{\tilde B-1}\nabla f(\boldsymbol{\hat u}_{j-\tau_{j,m},m,b})||^2\\\nonumber&+\frac{\mu(\mu^b-1)}{\mu-1}V^2+\frac{\mu(\mu^D-1)}{\mu-1}V^2\bigg)\bigg]+\frac{L\eta^2\rho_{t-1}^2}{2}M\tilde B V^2\\\nonumber &\leq -\frac{M\tilde B\eta\rho_{t-1}\alpha}{2}E[||\nabla f(\boldsymbol v_{t-1})||^2]+\frac{M\eta^3\rho_{t-1}\alpha L^2 V^2}{2}\sum_{b=0}^{\tilde B -1}\frac{\mu(\mu^b-1)}{\mu-1}\\\nonumber &+\frac{M\tilde B\eta^3L^2\rho_{t-1}\alpha V^2}{2}\frac{\mu(\mu^D-1)}{\mu-1}+\frac{L\eta^2\rho_{t-1}^2}{2}M\tilde B V^2+\frac{M^2\tilde B^2\eta^3L^2\rho_{t-1}\alpha}{2}\sigma^2\sum_{j=t-1-D'}^{t-2}\rho_{j-1}^2-\\\nonumber &\frac{\eta\rho_{t-1}\alpha}{2M\tilde B}E[||\sum_{m=1}^{M}\sum_{b=0}^{\tilde B-1}\nabla f(\boldsymbol{ \hat u}_{t-\tau_{t,m},m,b})||^2] +\frac{M\tilde B\eta^3L^2\rho_{t-1}\alpha D'}{2}\sum_{j=t-1-D'}^{t-2}\\ &\rho_{j-1}^2E||\sum_{m=1}^{M}\sum_{b=0}^{\tilde B-1}\nabla f(\boldsymbol{\hat u}_{j-\tau_{j,m},m,b})||^2
\end{align}
Summarising the inequality Eq.~\ref{proof_11}, from $t=1$ to $T$, we end up with:
\begin{align}\label{proof_12}
 \nonumber E[f(\boldsymbol v_{T})]-f(\boldsymbol v_{0}) &\leq -\frac{M\tilde B\eta\alpha}{2}\sum_{t=1}^{T}\rho_{t-1} E[||\nabla f(\boldsymbol v_{t-1})||^2]+\bigg(\sum_{b=0}^{\tilde B -1}\frac{\mu(\mu^b-1)}{\mu-1}+\tilde B\frac{\mu(\mu^D-1)}{\mu-1}\bigg)\frac{M\eta^3 L^2 V^2\alpha}{2}\sum_{t=1}^{T}\rho_{t-1}\\\nonumber &+\frac{L\eta^2}{2}M\tilde B V^2 \sum_{t=1}^{T}\rho_{t-1}^2+\frac{M^2\tilde B^2\eta^3L^2\alpha}{2}\sigma^2\sum_{t=1}^T\rho_{t-1}\sum_{j=t-1-D'}^{t-2}\rho_{j-1}^2-\frac{\eta\alpha}{2M\tilde B}\sum_{t=1}^T \rho_{t-1} E[||\sum_{m=1}^{M}\\\nonumber &\sum_{b=0}^{\tilde B-1}\nabla f(\boldsymbol{ \hat u}_{t-\tau_{t,m},m,b})||^2] +\frac{M\tilde B\eta^3L^2\alpha D'}{2}\sum_{t=1}^T\rho_{t-1}^2E||\sum_{m=1}^{M}\sum_{b=0}^{\tilde B-1}\nabla f(\boldsymbol{\hat u}_{t-\tau_{t,m},m,b})||^2\sum_{n=1}^{D'}\rho_{t+n}\\\nonumber &= -\frac{M\tilde B\eta\alpha}{2}\sum_{t=1}^{T}\rho_{t-1} E[||\nabla f(\boldsymbol v_{t-1})||^2]+\frac{M\eta^3 L^2 \alpha}{2}\sum_{t=1}^{T}\rho_{t-1}\bigg[V^2\bigg(\sum_{b=0}^{\tilde B -1}\frac{\mu(\mu^b-1)}{\mu-1}+\tilde B\frac{\mu(\mu^D-1)}{\mu-1}\bigg)\\\nonumber &+M\tilde B^2\sigma^2\sum_{j=t-1-D'}^{t-2}\rho_{j-1}^2\bigg]+\frac{L\eta^2}{2}M\tilde B V^2 \sum_{t=1}^{T}\rho_{t-1}^2 + \frac{\eta\alpha}{2}\sum_{t=1}^T \rho_{t-1}E[||\sum_{m=1}^{M}\sum_{b=0}^{\tilde B-1}\nabla f(\boldsymbol{ \hat u}_{t-\tau_{t,m},m,b})||^2]\bigg(\\&M\tilde B\eta^2L^2\rho_{t-1}D'\sum_{n=1}^{D'}\rho_{t+n}-\frac{1}{M\tilde B}\bigg)
\end{align}
Assuming that $\forall t\in\{1,2,...\}$:
\begin{equation}\label{proof_13}
M^2\tilde B^2\eta^2L^2\rho_{t-1}D'\sum_{n=1}^{D'}\rho_{t+n}\leq 1
\end{equation}
We have Theorem 1;
\begin{align}\label{proof_14}
\nonumber\sum_{t=1}^{T}\rho_{t-1} E[||\nabla f(\boldsymbol v_{t-1})||^2] &\leq \frac{2(f(\boldsymbol v_{0})-f(\boldsymbol v_{*}))}{ M\tilde B\eta\alpha} +\frac{\eta^2 L^2}{\tilde B} \sum_{t=1}^{T}\rho_{t-1}\bigg[V^2\bigg(\sum_{b=0}^{\tilde B -1}\frac{\mu(\mu^b-1)}{\mu-1}+\tilde B\frac{\mu(\mu^D-1)}{\mu-1}\bigg)\\ &+M\tilde B^2\sigma^2\sum_{j=t-1-D'}^{t-2}\rho_{j-1}^2\bigg]+\frac{L\eta V^2}{\alpha} \sum_{t=1}^{T}\rho_{t-1}^2
\end{align}

\textbf{Proof to Corollary 1:} According to Theorem 1, we have:
\begin{align}\label{proof_15}
\nonumber \frac{1}{\sum_{t=1}^{T}\rho_{t-1}}\sum_{t=1}^{T}\rho_{t-1} E[||\nabla f(\boldsymbol v_{t-1})||^2] &\leq \frac{2(f(\boldsymbol v_{0})-f(\boldsymbol v_{*}))}{ M\tilde B\eta\alpha\sum_{t=1}^{T}\rho_{t-1}} +\frac{\eta^2 L^2}{\tilde B\sum_{t=1}^{T}\rho_{t-1}} \sum_{t=1}^{T}\rho_{t-1}\bigg[V^2\bigg(\sum_{b=0}^{\tilde B -1}\frac{\mu(\mu^b-1)}{\mu-1}\\ &+\tilde B\frac{\mu(\mu^D-1)}{\mu-1}\bigg)+M\tilde B^2\sigma^2\sum_{j=t-1-D'}^{t-2}\rho_{j-1}^2\bigg]+\frac{L\eta V^2}{\alpha\sum_{t=1}^{T}\rho_{t-1}} \sum_{t=1}^{T}\rho_{t-1}^2
\end{align}
By setting the learning rate $\rho_t$ to be a constant and equal to $\eta$, we obtain the following:
\begin{equation}\label{proof_16}
 \frac{1}{T}\sum_{t=1}^{T} E[||\nabla f(\boldsymbol v_{t-1})||^2] \leq \frac{2(f(\boldsymbol v_{0})-E[f(\boldsymbol v_{*})])}{ TM\tilde B\eta^2\alpha} +\eta^2 LV^2\bigg(\frac{ 1}{\alpha}-\frac{L\mu(1-\mu^{\tilde B})}{(1-\mu)^2\tilde B}+ \frac{L\mu}{(1-\mu)}+ L\frac{\mu(1-\mu^D)}{1-\mu}\bigg)+\eta^4 L^2M\tilde B\sigma^2D'
\end{equation}
The inequalities shown in Theorem 1 can be re-arranged as follows:
\begin{equation}\label{proof_17}
\eta^4\leq \frac{1}{ M^2\tilde B^2L^2D'^2}
\end{equation}
Assuming $\eta\leq 1$ and $0<\mu< 1$,
\begin{equation}
\eta^4 \leq\frac{(\mu-1)^8}{\big(\mu(\mu-1)+9L^2\mu(D+1)(\mu^{D+1}-1)\big)^4}
\end{equation}
By setting the learning rate as follows:
\begin{equation}
    \eta^2=\frac{\sqrt{(f(\boldsymbol v_{0})-f(\boldsymbol v_{*}))}}{A\alpha \sqrt{TM\tilde B}}
\end{equation}
where $A$ is a constant:
\begin{equation}
\nonumber A=LV^2\bigg(\frac{1}{\alpha}+\frac{ 1}{\alpha^2}+ \frac{2L\mu}{(1-\mu)\alpha}\bigg)
\end{equation}
We end up with the following bound on the delayed parameters that if holds, Theorem 1 is satisfied:
\begin{equation}\label{proof_18}
T\geq \frac{M\tilde B L^2D'^2(f(\boldsymbol v_{0})-f(\boldsymbol v_{*}))}{A^2\alpha^2}
\end{equation}
\begin{equation}\label{proof_19}
T\geq \frac{\big(f(\boldsymbol v_{0})-f(\boldsymbol v_{*})\big)\big(\mu(\mu-1)+9L^2\mu(D+1)(\mu^{D+1}-1)\big)^4}{M\tilde BA^2\alpha^2(\mu-1)^8}
\end{equation}
Hence, from Theorem 1, we obtain:
\begin{align}\label{proof_20}
 \nonumber\frac{1}{T}\sum_{t=1}^{T} E[||\nabla f(\boldsymbol v_{t-1})||^2] &\leq \frac{2(f(\boldsymbol v_{0})-f(\boldsymbol v_{*}))}{ TM\tilde B\eta^2\alpha} +\eta^2 LV^2\bigg(\frac{ 1}{\alpha}+ \frac{2L\mu}{(1-\mu)}\bigg)+\eta^2 L\sigma^2\\\nonumber &\leq \frac{2(f(\boldsymbol v_{0})-f(\boldsymbol v_{*}))}{ TM\tilde B\eta^2\alpha} +\eta^2\alpha A\\&=3A\sqrt\frac{f(\boldsymbol v_{0})-f(\boldsymbol v_{*})}{TM\tilde B}
\end{align}
Therefore, Corollary 1 has been proven.

\section{Asynchronous Distributed Lock-free Parallel Stochastic Variational Inference}\label{applica}

In this section, we describe our proposed distributed parallel implementation of the asynchronous lock-free stochastic variational inference algorithm (DPSVI) on a \textit{computer cluster} with multi-core nodes. The steps of the algorithm follow from the proposed DPSGD but in the context of VI. First, we derive the model family applicable with DPSVI and review SVI following the same steps in~\citep{hoffman2013stochastic}. Then, we derive DPSVI from DPSGD. 

\textbf{Model family:} The family of models considered here consists of three random variables: observations $\boldsymbol x=\boldsymbol{x}_{1:n}$, local hidden variables $\boldsymbol z=\boldsymbol{z}_{1:n}$, global hidden variables $\boldsymbol\beta$ and fixed parameters $\boldsymbol\zeta$. The model assumes that the distribution of the $n$ pairs of $(\boldsymbol{x}_i,\boldsymbol{z}_i)$ is conditionally independent given $\boldsymbol\beta$. Furthermore, their distribution and the prior distribution of $\boldsymbol\beta$ are in the exponential family.
\begin{equation}\label{equ1}
p(\boldsymbol\beta,\boldsymbol x,\boldsymbol z|\boldsymbol\zeta)=p(\boldsymbol\beta|\boldsymbol\zeta)\prod_{i=1}^n p(\boldsymbol{z}_i,\boldsymbol{x}_i|\boldsymbol\beta)
\end{equation}
\begin{equation}\label{equ2}
p(\boldsymbol{z}_i,\boldsymbol{x}_i|\boldsymbol\beta)=h(\boldsymbol{x}_i,\boldsymbol{z}_i)\exp\big(\boldsymbol\beta^Tt(\boldsymbol x_i,\boldsymbol z_i)-a(\boldsymbol\beta)\big)
\end{equation}
\begin{equation}\label{equ3}
p(\boldsymbol\beta|\boldsymbol\zeta)=h(\boldsymbol\beta)\exp\big(\boldsymbol\zeta^Tt(\boldsymbol\beta)-a(\boldsymbol\zeta) \big)
\end{equation}
Here, we overload the notation for the base measures $h(.)$, sufficient statistics $t(.)$ and log normaliser $a(.)$. While the proposed  approach is generic, for simplicity we assume a conjugacy relationship between $(\boldsymbol x_i,\boldsymbol z_i)$ and $\boldsymbol\beta$. That is, the distribution $p(\boldsymbol\beta|\boldsymbol x,\boldsymbol z)$ is in the same family as the prior $p(\boldsymbol\beta|\boldsymbol\zeta)$. Note that this family of models includes, but is not limited to, latent Dirichlet allocation, Bayesian Gaussian mixture, probabilistic matrix factorisation, hidden Markov models, hierarchical  linear and probit regression and many Bayesian non-parametric models.  

\textbf{Mean-field variational inference.} Variational inference (VI) approximates intractable posterior $p(\boldsymbol\beta,\boldsymbol z|\boldsymbol x)$ by positing a family of simple distributions $q(\boldsymbol\beta,\boldsymbol z)$ and find the member of the family that is closest to the posterior (closeness is measured with KL divergence). The resulting optimisation problem is equivalent to maximising the evidence lower bound (ELBO).
\begin{equation}\label{equ4}
\mathcal{L}(q)=E_q[\log p(\boldsymbol x,\boldsymbol z,\boldsymbol\beta)]-E_q[\log p(\boldsymbol z\boldsymbol\beta)]\leq \log p(\boldsymbol x)
\end{equation}
 Mean-field is the simplest family of distribution, where the distribution over the hidden variables factorises as follows:
 \begin{equation}\label{equ5}
 q(\boldsymbol\beta,\boldsymbol z)=q(\boldsymbol\beta|\boldsymbol\lambda)\prod_{i=1}^np(\boldsymbol z_i|\boldsymbol\phi_i)
 \end{equation}
 Further, each variational distribution is assumed to come from the same family of the true one. Mean-field variational inference optimises the new ELBO with respect to the  local and global variational parameters $\boldsymbol\phi$ and $\boldsymbol\lambda$.
 \begin{equation}\label{equ6}
\mathcal{L}(\boldsymbol\lambda,\boldsymbol\phi)=E_q\bigg[\log\frac{p(\boldsymbol \beta)}{q(\boldsymbol \beta)}\bigg]+\sum_{i=1}^nE_q\bigg[\log\frac{p(\boldsymbol x_i,\boldsymbol z_i|\boldsymbol\beta)}{q(\boldsymbol z_i)} \bigg]
 \end{equation}
  It iteratively updates each variational parameter holding the other parameters fixed. With the assumptions taken so far, each update has  a closed form solution. The local parameters are a function of the global parameters.
 \begin{equation}\label{equ7}
 \boldsymbol\phi({\boldsymbol\lambda}_t)=\arg\max_{\boldsymbol\phi}\mathcal{L}(\boldsymbol\lambda_t,\boldsymbol\phi)
\end{equation}  
  We are interested in the global parameters which summarise the whole dataset (clusters in the Bayesian Gaussian mixture, topics in LDA).
 \begin{equation}\label{equ8}
 \mathcal{L}(\boldsymbol\lambda)=\max_{\boldsymbol\phi} \mathcal{L}(\boldsymbol\lambda,\boldsymbol\phi)
\end{equation}  
To find the optimal value of $\boldsymbol\lambda$ given that $\boldsymbol\phi$ is fixed, we compute the natural gradient of $\mathcal{L}(\boldsymbol\lambda)$  and set it to zero by setting:
 \begin{equation}\label{equ9}
\boldsymbol\lambda^*=\boldsymbol\zeta +\sum_{i=1}^nE_{\boldsymbol\phi_i({\boldsymbol\lambda}_t)}[t(\boldsymbol x_i,\boldsymbol z_i)]
 \end{equation} 
Thus, the new optimal global parameters are $\boldsymbol\lambda_{t+1}=\boldsymbol\lambda^*$.  The algorithm works by iterating between computing  the optimal local parameters  given the global ones \big(Eq.~\ref{equ7}\big) and computing the optimal global parameters given the local ones \big(Eq.~\ref{equ9}\big).
 
\textbf{Stochastic variational inference.} Instead of analysing all of the data to compute $\boldsymbol\lambda^*$ at each iteration, stochastic optimisation can be used. Assuming that the data samples are uniformly randomly selected from the dataset, an unbiased noisy estimator of $\mathcal{L}(\boldsymbol\lambda,\boldsymbol\phi)$  can be developed based on a single data point: 
\begin{equation}\label{equ10}
\mathcal{L}_i(\boldsymbol\lambda,\boldsymbol\phi_i)=E_{q}\bigg[\log\frac{p(\boldsymbol \beta)}{q(\boldsymbol \beta)}\bigg]+nE_q\bigg[\log\frac{p(\boldsymbol x_i,\boldsymbol z_i|\boldsymbol\beta)}{q(\boldsymbol z_i)} \bigg]
\end{equation}\label{equ11}
The unbiased stochastic approximation of the ELBO as a function of $\boldsymbol\lambda$ can be written as follows:
\begin{equation}\label{equ11b}
\mathcal{L}_i(\boldsymbol\lambda)=\max_{\boldsymbol\phi_i}\mathcal{L}_i(\boldsymbol\lambda,\boldsymbol\phi_i)
\end{equation}
Following the same steps in the previous section, we end up with a noisy unbiased estimate of Eq.~\ref{equ8}: 
\begin{equation}\label{equ12}
\boldsymbol{\hat{\lambda}}=\boldsymbol\zeta +nE_{\boldsymbol\phi_i({\boldsymbol\lambda}_t)}[t(\boldsymbol x_i,\boldsymbol bz_i)]
\end{equation}
At each iteration, we move the global parameters a step-size $\rho_t$ (learning rate) in the direction of the noisy natural gradient:
 \begin{equation}\label{equ13}
\boldsymbol\lambda_{t+1}=(1-\rho_t)\boldsymbol\lambda_t+\rho_t\boldsymbol{\hat{\lambda}}
\end{equation}
With certain conditions on $\rho_t$, the algorithm converges ($\sum_{t=1}^\infty\rho_t=\infty$, $\sum_{t=1}^\infty \rho_t^2<\infty$)~\citep{robbins1951stochastic}.

Based on a batch of data points, the unbiased noisy estimator of $\mathcal{L}(\boldsymbol\lambda,\boldsymbol\phi)$ can be written as follows:
\begin{equation}\label{Intro_eq2}
\mathcal{L}_g(\boldsymbol\lambda,\boldsymbol\phi_g)=E_{q}\bigg[\log\frac{p(\boldsymbol \beta)}{q(\boldsymbol \beta)}\bigg]+\frac{n}{G}\sum_{i\in G_g} E_q\bigg[\log\frac{p(\boldsymbol x_i,\boldsymbol z_i|\boldsymbol\beta)}{q(\boldsymbol z_i)} \bigg]
 \end{equation}
 where $G_g=\{(g-1)G+1,...,gG\}$. Equation~\ref{equ12} can be written as follows:
 
 \begin{equation}\label{equ14}
\boldsymbol{\hat{\lambda}_g}=\boldsymbol\zeta +\frac{n}{G}\sum_{i\in G_g}E_{\boldsymbol\phi_i({\boldsymbol\lambda}_t)}[t(\boldsymbol x_i,\boldsymbol z_i)]
\end{equation}
Following this derivation, DPSVI, Alg.~\ref{alg31} and \ref{alg32}, can be simply obtained from DPSGD by amending line (7) and (8) of Alg.~\ref{alg22}. Specifically, in line (7) we compute the local variational parameters $\boldsymbol\phi_i({{\boldsymbol\lambda^*}})$ corresponding to the  data point $\boldsymbol x_i$ and the global variational parameter ${\boldsymbol \lambda^*}$, $\boldsymbol\phi_i({\boldsymbol\lambda^*})=\arg\max_{\boldsymbol\phi_i}\mathcal{L}_i({\boldsymbol\lambda^*},\boldsymbol\phi_i)$. The index $i$ is randomly picked from $\{1,... n\}$ and  ${\boldsymbol \lambda^*}$ replaces $\hat{\boldsymbol u}$. In line (8) we perform the update after replacing the stochastic gradient $\nabla f_i(\hat{\boldsymbol u})$ by negative stochastic natural gradient with respect to the global parameter ${\boldsymbol\lambda^*}$, $\boldsymbol g_i({\boldsymbol\lambda^*})=-\bigg(\boldsymbol\zeta +\frac{n}{G}\sum_{j\in G_i}E_{\boldsymbol\phi_j({{\boldsymbol\lambda^*}})}[t(\boldsymbol x_j,\boldsymbol z_j)]-{\boldsymbol\lambda^*}$\bigg). 

Finally, the derivation of LDA from the presented model family can be found in~\citep{smohamad}.

\begin{algorithm}[t]
   \caption{DPSVI-Master: Updates performed at the master machine}\label{alg31}
\begin{algorithmic}[1]
       \STATE \textbf{initialise:} number of iteration $T$, global variable $\boldsymbol v$, global learning rate $\{\rho_t\}_{t=0,...,T-1}$ 
       \FOR{$t=0,1,2,...T-1$}
        \STATE Collect $M$ updating vectors $\boldsymbol w_1,...,\boldsymbol w_M$ from the workers.
      \STATE Update the current estimate of the global parameter $\boldsymbol v \leftarrow \boldsymbol v+\rho_{t}\sum_m \boldsymbol w_m$
      \STATE $t\leftarrow t+1$
     \ENDFOR
\end{algorithmic}
\end{algorithm}
\begin{algorithm}[t]
   \caption{DPVI-Worker: Updates performed at each worker machine}\label{alg32}
\begin{algorithmic}[1]
  \STATE \textbf{initialise:} number of iterations of per-worker loop $B$, learning rate $\eta$, number of threads $p$ 
   \WHILE{(MasterIsRun)}
        \STATE Pull a global parameter $\boldsymbol v$ from the master and put it in the shared memory.
        \STATE Fork p threads
        \FOR{c=0 to B-1}
            \STATE Read current values of $\boldsymbol \lambda^*$, denoted as $\boldsymbol{\hat  \lambda^*}$, from the shared memory.
        \STATE Randomly pick i from $\{1,...n\}$
    \STATE Compute the local variational parameters $\boldsymbol\phi_i^*(\boldsymbol{\hat  \lambda^*})$ corresponding to the  data point $\boldsymbol x_i$ and the global variational parameter $\boldsymbol{\hat  \lambda^*}$, $\boldsymbol\phi_i^*(\boldsymbol{\hat  \lambda^*})=\arg\max_{\boldsymbol\phi_i}\mathcal{L}_i(\boldsymbol{\hat  \lambda^*},\boldsymbol\phi_i)$
    \STATE  Compute the stochastic natural gradient with respect to the global parameter $\boldsymbol\lambda$, $\boldsymbol g_i(\boldsymbol\lambda,\boldsymbol\phi)=-\bigg(\boldsymbol\zeta +\frac{n}{G}\sum_{j\in G_i}E_{\boldsymbol\phi_j({{\boldsymbol\lambda}})}[t(\boldsymbol x_j,\boldsymbol z_j)]-{\boldsymbol\lambda}\bigg)$
        \STATE Update the local version of the global variational parameter $\boldsymbol{ \lambda^*}\leftarrow \boldsymbol{ \lambda^*} +\eta g_i(\boldsymbol{\hat  \lambda^*,\boldsymbol\phi^*})$
     \ENDFOR
     \STATE Push the update vector $\boldsymbol{ \lambda^*}-\boldsymbol{v}$ from the shared memory to the master
     \ENDWHILE
\end{algorithmic}
\end{algorithm}

\section{Highly Scalable Advantage Actor Critics}\label{Drl_n}
The impressive results that have been achieved by deep artificial neural networks in several application domains are often driven by the availability of very large training data sets~\citep{krizhevsky2012imagenet}. In reinforcement learning (RL)~\citep{sutton1998introduction}, an agent learns how to behave by interacting with its environment and has to experiment by trial and error, over and over again, often accumulating millions of repeated experiences. In order to enable learning in complex, real-world environments, recent advances in RL have successfully incorporated function approximation through deep networks resulting in deep reinforcement learning (DRL)~\citep{sutton1998introduction}. The already data hungry DL function is then  aggravated  by the data inefficiency of RL motivating the development of more scalable learning algorithms.

Policy gradient methods directly maximise the expected rewards of a parameterised policy using gradient-based iterative methods such as the Stochastic Gradient Descent method (SGD). Advantage actor-critic incorporates control variate techniques to reduce the variance of the approximated gradient~\citep{sutton2000policy}. Thus, different versions of SGD can be used directly for learning. Various SGD implementations have been exploited to scale up DRL including distributed~\citep{agarwal2011distributed,lian2015asynchronous} and parallel algorithms~\citep{recht2011hogwild,zhao2017lock} resulting in different scalable DRL algorithms~\citep{ong2015distributed,nair2015massively,adamski2018distributed,mnih2016asynchronous,babaeizadeh2016reinforcement,clemente2017efficient,horgan2018distributed}. These developments are particularly relevant to DRL due to its inherently sequential nature and the massive amount of data required for learning complex tasks such as playing games~\citep{silver2016mastering}, controlling robots~\citep{abbeel2007application}, optimising memory control~\citep{ipek2008self}, and personalising web services~\citep{theocharous2015personalized}, amongst others.

More recently, a few hybrid DRL algorithms have been proposed that combine both aspects of parallel (i.e. shared) and distributed computation (i.e., distributed memory)~\citep{adamski2018distributed,babaeizadeh2016reinforcement}. A typical problem of distributed learning is the communication overhead arising from the necessity to share the weight updates between nodes. Traditionally, large batches and step sizes have been used to curb the communication while preserving scalability. These strategies introduce a trade-off between computational and sample efficiency: a large batch increases the time needed to calculate the gradient locally, but decreases its variance allowing higher learning rate to be used. However, there is a limit on the speed-up that can be achieved by tuning the learning rate and batch sizes~\citep{li2014efficient,bottou2018optimization}. 

Here, we apply  DPSGD to steer off communication computation in a hybrid distributed-parallel implementation of DRL, with initial focus on advantage actor critic (A2C), which has been extensively studied~\citep{ong2015distributed,adamski2018distributed,mnih2016asynchronous,babaeizadeh2016reinforcement,clemente2017efficient,horgan2018distributed}. We use a cluster network with a given number of computational nodes equipped with multiple CPUs. Each node maintains a copy of the actor and critic's models; for instance, the weights of the corresponding neural network implementing those models. A few of these nodes act as a {\it master} and the remaining nodes are the {\it workers}. The models maintained by the master are called global models. Within each worker node, the A2C's models are shared locally among its multiple CPUs, and are denoted as {\it local}. 

Each worker performs multiple lock-free parallel updates~\citep{zhao2016fast} for the models shared across the CPUs. The global model is then updated by the master using the asynchronously aggregated local multiple steps updates. This simple strategy yields a highly-scalable advantage actor-critics (HSA2C), and harnesses distributed and local computation and storage resources.  By updating the local variables multiple times, HSA2C mitigates the communication cost converting time speed-up. 

\subsection{Background}\label{app_sec2}
 In reinforcement learning (RL), an agent interacts  sequentially with an environment, with the goal of maximising cumulative rewards.  At each step $t$ the agent observes a state $s_t$, selects an action $a_t$ according to its policy $\pi(a_t|s_t)$, and receives the next state $s_{t+1}$ along with a reward $r_t$. This continues until the agent reaches a terminal state at $t=T$. The cumulative rewards, called return, at time $t$ can be then written as $R_t=\sum_{i=0}^{\infty}\gamma^i r_{t+i}$, where the goal is to learn a policy that maximises the expected return from each state $s_t$ : $E[R_t|s_t=s]$. The action value function of policy $\pi$,$Q^{\pi}(s,a)=E[R_t|s_t=s,a_t=a,\pi]$ is the expected return for taking action $a$ in state $s$ and following policy $\pi$. The value function $V^{\pi}(s)=E[R_t|s_t=s]$ is the expected return of policy $\pi$ from state $s$.
 
 Two main approaches of RL have been studied: value-based and policy based RL. In value-based RL, the policy is inferred from the value function which is represented by a function approximator such as a neural network. Hence, the value function can be written as $Q(s,a; w)$, where $w$ represents the approximator parameters. Then, the goal of the RL algorithms is to iteratively update $w$ to find the optimal action value function representing the optimal policy $\pi$. Alternatively, policy based DRL directly parameterises the policy $\pi(\bm{a}|\bm{s},\bm{\theta})$ and updates its parameter by performing, typically approximate, gradient ascent on 
$$
L(\bm{\theta})=E[Q^{\pi_{\bm{\theta}}}(\bm{a},\bm{s})|\bm{\theta}]
$$
e.g. see~\citep{williams1992simple}. Hence, the gradient of the objective function can be expressed as follows:
\begin{equation}\label{Equ3}
  \nabla L(\bm{\theta})=E[\nabla_{\bm{\theta}}\log\pi(\bm{a}|\bm{s},\bm{\theta})Q^{\pi_{\bm{\theta}}}(\bm{a},\bm{s})|\bm{\theta}]
\end{equation}
An unbiased estimate of the gradient in Eq.~\ref{Equ3} can be obtained by computing the update from randomly sampled tuples of form $(\bm{s},\bm{a},r)$. 

\subsubsection{Advantage actor critic (A2C) methods}
To reduce the variance of the estimate in Eq.~\eqref{Equ3} while keeping it unbiased, the $Q$ function can be replaced with an advantage function, $A^{\pi_{\bm{\theta}}}(\bm{a},\bm{s})=(Q^{\pi_{\bm{\theta}}}(\bm{a},\bm{s})-V^{\pi_{\bm{\theta}}}(\bm{s}))$~\citep{sutton2000policy} where $V$ is the state value function. This approach can be viewed as an actor-critic architecture where the policy is the actor and the advantage function is the critic~\citep{sutton1998reinforcement}.

Deep neural networks are used to approximate the actor and critics functions. Typically, two neural networks are deployed, one parameterised by $\bm{\theta}$ approximating the actor and the other parameterised by $\bm{\theta}_v$ approximating the critic. Hence, the gradient of the objective function can be expressed as follows:

\begin{equation}\label{Equ332}
  \nabla L(\bm{\theta},\bm{\theta}_v)=\begin{pmatrix} \nabla_{\bm{\theta}} L(\bm{\theta},\bm{\theta}_v) \\ \nabla_{\bm{\theta}_v} L(\bm{\theta},\bm{\theta}_v) \end{pmatrix}
  \end{equation}
  where 
  \begin{align}\label{Equ333}
  \nonumber\nabla_{\bm{\theta}} L(\bm{\theta},\bm{\theta}_v)=&E[\nabla_{\bm{\theta}}\log\pi(\bm{a}|\bm{s},\bm{\theta})A^{{\bm{\theta}_v}}(\bm{a},\bm{s})|\bm{\theta}]\\
  \nabla_{\bm{\theta}_v} L(\bm{\theta},\bm{\theta}_v)=&E[\nabla_{\bm{\theta}_v}A^{{\bm{\theta}_v}}(\bm{a},\bm{s})|\bm{\theta}]
\end{align}

In the next section we provide a brief overview of existing scalable SGD approaches and how they have been adopted to scale-up A2C.

\subsubsection{Scalable SGD algorithms for DRL}\label{app_subsec3}

Stochastic gradient descent (SGD) and its variants are used to optimise the A2C objective function (Eq.~\eqref{Equ332})~\citep{sutton1998introduction}. SGD updates the actor and the critics network weights $(\bm\theta,\bm\theta_v)$ based on the approximate gradient of the objective function $L(\bm\theta,\bm\theta_v)$ computed using an experience trajectory (or batch of trajectories) sampled using policy $\pi_{\theta}$. Various scalable SGD-based approaches have recently been proposed to scale up DRL algorithms~\citep{ong2015distributed,nair2015massively,mnih2016asynchronous,babaeizadeh2016reinforcement,clemente2017efficient,horgan2018distributed,adamski2018distributed}. In general, these approaches can be viewed as derivations from either distributed SGD (DSGD) or parallel SGD (PSGD).

A DSGD-like architecture called Gorila,  proposed by~\cite{nair2015massively}, relies on asynchronous training of RL agents in a distributed setting. Gorila distributes Deep Q-Network DQN~\citep{mnih2015human} across multiple machines. Each machine runs an actor that interacts with the environment, samples data from the replay memory and computes the gradients of the DQN loss with respect to the policy parameters. The gradients are asynchronously sent to a central parameter server which updates a central copy of the model. The updated policy parameters are sent to the actor-learners at fixed intervals. In~\citep{mnih2016asynchronous}, PSGD-like parameter updates and data generation have been used within a single-machine, in a multi-threaded rather than a distributed context. The shared parameter is then updated in an asynchronous DSGD-like fashion. 
\begin{algorithm}[ht]
   \caption{PA3C}\label{app_alg1}
\begin{algorithmic}[1]
  \STATE \textbf{initialise:} number of threads $p$, thread step counter $t\leftarrow 0$,iteration counter $T\leftarrow 0$, thread gradient accumulator $\bm{g}\leftarrow 0$ and $\bm{g}_v\leftarrow 0$, shared update vector accumulator $\bm{G}\leftarrow 0$ and $\bm{G}_v\leftarrow 0$, mini-batch size $m$, mini-batch counter $n$, shared parameter vector $\bm \theta$ and $\bm \theta_v$,  learning rate.
  \STATE Fork p threads
  \REPEAT
      \STATE $t_{start}=t$
      \STATE Get state $\bm{s_t}$
        \REPEAT
            \STATE Read current values of $\bm \theta$ , denoted as $\bm{\hat \theta}$, from the shared memory.
            \STATE Take action $\bm{a_t}$ according to policy $\pi(\bm{a}|\bm{s_t};\bm{\hat\theta})$
            \STATE  Receive new state $\bm{s_{t+1}}$ and reward $r_t$
            \STATE $t\leftarrow t+1$
            \UNTIL{terminal $\bm{s_t}$ or $t-t_{start}==t_{max}$}
            \STATE  Read current values of $\bm{\theta_v}$ , denoted as $\bm{\hat \theta_v}$, from the shared memory.
            \STATE R=$\begin{cases} 0 & \text{for terminal } \bm{s_t}\\  V(\bm{s_t},\bm{\hat \theta_v}) &\text{for non-terminal } \bm{s_t} \end{cases}$
            \FOR{$i\in\{t-1,...,t_{start}\}$}
            \STATE $R\leftarrow r_i+\gamma R$
            \STATE Accumulate gradients wrt $\bm{\hat \theta}$: $\bm{g}\leftarrow \bm{g}+\nabla_{\bm{\hat \theta}}\log \pi(\bm{a_i}|\bm{s_i};\bm{\hat{\theta}})(R-V(\bm{s_i;\bm{\hat{\theta}_v}}))$
            \STATE Accumulate gradients wrt $\bm{\hat \theta_v}$: $\bm{g_v}\leftarrow \bm{g_v}+\frac{\partial(R-V(\bm{s_i;\bm{\hat{\theta}_v}}))^2}{\partial \bm{\hat \theta_v}}$
            \ENDFOR
        \STATE Accumulate mini-batch gradients: $\bm{G}\leftarrow \bm{G}+\bm{g}$ and $\bm{G}_v\leftarrow\bm{G}_v+ \bm{g_v}$
    \STATE $\bm g\leftarrow 0$
      \STATE $\bm{g_v}\leftarrow 0$
              \STATE $n\leftarrow n+1$
         \IF{$n==m$}
         \STATE $n=0$
         \STATE Update the shared network weights $\bm{ \theta}$ using $\bm{G}$ and the learning rate
      \STATE Update the shared network weights $\bm{ \theta_v}$ using $\bm{G_v}$ and the learning rate
    \STATE$\bm{G}=0$ and $\bm{G}_v=0$
      \STATE Update local steps: $T\leftarrow T+1$
      \ENDIF
    \UNTIL{$T>T_{max}$}
\end{algorithmic}
\end{algorithm}
Scaling-up A2C using this two architectures lead to the asynchronous distributed A2C (DA3C) in Alg.~\ref{app_alg11} and Alg.~\ref{app_alg12} and the lock-free parallel A2C (PA3C) in Alg.~\ref{app_alg1}. The lock-free PSGD presented in Alg.~\ref{app_alg1} employs~\citep{recht2011hogwild} style of updates for training A2C. The actors and critics networks are stored in a shared memory where different threads can update their parameters without any memory locking. Therefore, in theory, linear speed-ups with respect to the number of threads can be achieved. On the other hand, the asynchronous DSGD presented in Alg.~\ref{app_alg11} and Alg.~\ref{app_alg12} employs~\citep{agarwal2011distributed} style of updates for training A2C. Here, the algorithms are deployed on a cluster where a master machine maintains a copy of A2C's networks (the global networks $(\bm{\theta},\bm{\theta_v})$, see Alg.~\ref{app_alg11}). Other machines serve as workers which independently and simultaneously compute the local stochastic gradients of a copy of A2C's networks (the local networks, see Alg.~\ref{app_alg12}). The workers only communicate with the master to exchange information in which they access the state of the global networks and provide the master with the stochastic gradients. The master aggregates predefined amounts of gradients from the workers. Then, it updates its global networks. Note that the local distributed computations are done in an asynchronous style where the workers are not locked until the master starts updating the global networks. That is, the workers might compute some stochastic gradients based on early value of the global networks.

\begin{algorithm}[ht]
   \caption{DA3C-Master}\label{app_alg11}
\begin{algorithmic}[1]
       \STATE \textbf{initialise:} global network parameters $\bm{\theta}$ and $\bm{\theta_v}$, batch size $M$, global learning rate
       \FOR{$t=0,1,2,...T-1$}
        \STATE Collect $M$ updating vectors $(\bm w_1,\bm w_{v1}),...,(\bm w_M,\bm w_{vM})$ from the workers.
      \STATE Update the current estimate of the global parameters $(\bm{\theta},\bm{\theta_v})$ using $\sum_{m=1}^M (\bm w_m,\bm w_{vm})$
     \ENDFOR
\end{algorithmic}
\end{algorithm}

\begin{algorithm}[ht]
   \caption{DA3C-Worker}\label{app_alg12}
\begin{algorithmic}[1]
  \STATE \textbf{initialise:} batch counter $n\leftarrow 0$, counter $t\leftarrow 0$, gradient accumulator $\bm{g}\leftarrow 0$ and $\bm{g}_v\leftarrow 0$, gradient vector $\bm{G}\leftarrow 0$ and $\bm{G}_v\leftarrow 0$, mini-batch size $m$, local networks parameters $\bm \theta$ and $\bm \theta_v$
  \REPEAT
   \STATE Pull global networks from the master and assign them to $(\bm \theta,\bm \theta_v)$ 
      \STATE $t_{start}=t$
      \STATE Get state $\bm{s_t}$
        \REPEAT
            \STATE Take action $\bm{a_t}$ according to policy $\pi(\bm{a}|\bm{s_t};\bm{\theta})$
            \STATE  Receive new state $\bm{s_{t+1}}$ and reward $r_t$
            \STATE $t\leftarrow t+1$
            \UNTIL{terminal $\bm{s_t}$ or $t-t_{start}==t_{max}$}
            \STATE R=$\begin{cases} 0 & \text{for terminal } \bm{s_t}\\  V(\bm{s_t},\bm{\hat \theta_v}) &\text{for non-terminal } \bm{s_t} \end{cases}$
            \FOR{$i\in\{t-1,...,t_{start}\}$}
            \STATE $R\leftarrow r_i+\gamma R$
            \STATE Accumulate gradients wrt $\bm{ \theta}$: $\bm{g}\leftarrow \bm{g}+\nabla_{\bm{\theta}}\log \pi(\bm{a_i}|\bm{s_i};\bm{{\theta}})(R-V(\bm{s_i;\bm{{\theta}_v}}))$
            \STATE Accumulate gradients wrt $\bm{\theta_v}$: $\bm{g_v}\leftarrow \bm{g_v}+\frac{\partial(R-V(\bm{s_i;\bm{{\theta}_v}}))^2}{\partial \bm{\theta_v}}$
            \ENDFOR
        \STATE Accumulate mini-batch gradients: $\bm{G}\leftarrow \bm{G}+\bm{g}$ and $\bm{G}_v\leftarrow\bm{G}_v+ \bm{g_v}$
    \STATE $\bm g\leftarrow 0$
      \STATE $\bm{g_v}\leftarrow 0$
              \STATE $n\leftarrow n+1$
         \IF{$n==m$}
         \STATE $n=0$
       \STATE Push the gradient vectors $\bm{G}$ and $\bm{G}_v$ to the master
    \STATE$\bm{G}=0$ and $\bm{G}_v=0$
      \ENDIF
    \UNTIL{Master Stop}
\end{algorithmic}
\end{algorithm}

Inspired by these two approaches, different improvements have been proposed. In~\citep{babaeizadeh2016reinforcement}, a hybrid CPU/GPU version of the Asynchronous Advantage Actor Critic (A3C) algorithm~\citep{mnih2016asynchronous} was introduced. This study focused on mitigating the severe under-utilization of the GPU computational resources in DRL caused by its sequential nature of data generation. In this work, each agent queues policy requests in a \textit{Prediction Queue} before each action, and periodically submits a batch of reward experiences to a \textit{Training Queue}. Thus, unlike~\citep{mnih2016asynchronous}, the agents do not compute the gradients themselves. Instead, they send experiences to central learners that update the network on the GPU accordingly. Such architecture reduces the GPU idle during the training. However, as the number of core increases, the leaner becomes unable to cope with the data. Furthermore, such amount of data requires large storage capacity. Besides, the internal communications can affect the speed-up when bandwidth reach its ceiling. These issues limit the scalability of ~\citep{babaeizadeh2016reinforcement}. The approach is also limited to Off-policy DRL methods. We also note that a similar way for parallelisation of DRL is proposed by~\cite{clemente2017efficient}. 

Similarly, \cite{horgan2018distributed} proposes to generate experience data in parallel using multi-cores CPUs where experiences are accumulated by actors in a shared experience replay memory. Each actor interacts with the environment takes actions, gets rewards and states. The learner, then, replays samples of experience and updates the shared neural network. The architecture relies on prioritised experience replay~\citep{schaul2015prioritized} to focus only on the most significant data generated by the actors. Along the same trend, in \citep{espeholt2018impala}, all the data (states, actions and rewards) is also accumulated by distributed actors and communicated to a centralised learner where the computation takes place. To correct the policy-lag between the learner and actors generating the data, these authors introduce the V-trace off-policy actor-critic algorithm. It is an importance weighting technique used to learn the target policy (learner policy) from behaviour ones (actors policy). The architecture of these studies~\citep{horgan2018distributed,espeholt2018impala} allows distributing the generation and selection of experience data instead of distributing locally computed gradients as in~\citep{nair2015massively}. Hence, it requires sending large size information over the network in case of large size batch of data making the communications more problematic.  Furthermore, the central learner has to perform  most of the computation which limits the scalability. Moreover, these approach is limited to Off-policy DRL methods.

The work in ~\citep{adamski2018distributed} presents an SGD-based hybrid distributed-parallel actor critic approach, and is the most closely related to our approach. These authors combine PA3C with DA3C to allow parallel distributed implementation of A3C on a computer cluster with multi-core node. Each node applies~\citep{babaeizadeh2016reinforcement} to queue data in batches, which are used to compute local gradients. These gradients are gathered from all workers, averaged and applied to update the global network parameters. To reduce the communication overhead, a careful reexamination of Adam optimizer’s hyper-parameters is carried out allowing large batch sizes to be used.


\subsection{Scalable actor-critic through multiple local updates}\label{app_sec3}

We propose a hybrid distributed-parallel actor critic algorithm whereby, unlike~\citep{adamski2018distributed}, workers communicate update vectors rather than gradients to the master. These update vectors are the discrepancy between the initial copy of the advantage actor critic (A2C)'s local network parameters pulled from the master and the results of multiple local iterations performed over the networks using PSGD-like updating style. As in DA3C-like algorithms, the master aggregates predefined amounts of these local updates, updates its global networks and broadcast the updated networks to the workers. Such multi-steps updating vectors allow large global update steps with less communication exchanges. The local A2C's actor model keeps interacting with the environment, taking actions and receiving rewards and next states. The data being collected is constantly used to compute the gradient of the objective function of A2C's actor and critic networks and updating these networks. The data collection and updates are done in PA3C-like style. Once a pre-defined number of updates is reached the local update vector is computed and pushed to the master in DA3C style preventing any locking or synchronisation (unlike~\citep{adamski2018distributed}) from halting the actor infraction with the environment. 

One issue of this approach is that the variance of policy gradient DRL combinatorially increases with each step because of their Markovian dependency. Hence, there is a trade-off between variance and communication cost. Higher number of multi-step update results in higher variance, but reduces the communication cost. In our approach, the variance issue is slightly mitigated by the control variate technique used by the A2C algorithm. We also adopt a local mini-batch along with multi-step TD error to further reduce the variance. The variance can also be reduced when increasing the global batch size (i.e., the number of aggregated update vectors). Thus, adding more nodes reduces the variance allowing a higher number of multi-step updates. As the number of nodes increases, communication exchanges also increase. Therefore, our approach is beneficial when the number of nodes increases, as this involves increased communication, up to a certain breaking point after which the higher variance associated with higher number of multi-step updates counterbalances the benefits.
\subsubsection{Overview of the algorithm}
HSA2C is presented in Alg.~\ref{alg51} and ~\ref{alg52}. On a single node agents work in parallel without any locking for the shared memory. Each agent interacts with an instance of the game environment producing experience data (state, action and reward) (line 8 to 10 in Alg.~\ref{alg52}). HSA2C uses $k$ steps TD error along with the function approximation to allow better bias-variance trade-off (line 14 to 19 in Alg.~\ref{alg52}). Since the steps are taken in parallel by multiple independent actors, data correlation is reduced resulting in lower bias without the need for experience replay making the proposed approach applicable for both off-policy and on-policy DRL methods~\citep{mnih2016asynchronous}. HSA2C uses local mini-batch (line 20 to line 24 in Alg.~\ref{alg52}) to reduce the stochastic gradient variance which is crucial to achieve the speed-up. That is HSA2C uses mini-batch of stochastic gradients, computed in parallel from different experiences, to estimate lower variance gradient. Such  gradients allow more stable multiple updates (line 24 to 29 in Alg.~\ref{alg52}). Finally, HSA2C's workers send the computed multiple steps update vectors (in lock-free parallel) to the master (line 30 to 32 in Alg~\ref{alg52}) which asynchronously collect these vectors and update its global parameters (Alg~\ref{alg51}). The workers then pull the update global parameters and update  their local parameters (line 33 to 34 in Alg~\ref{alg52}).

The convergence rate for serial and synchronous parallel stochastic gradient (SG) is consistent with $O(1/\sqrt{T})$~\citep{ghadimi2013stochastic,dekel2012optimal,nemirovski2009robust}. As per corollary 1, HSA2C achieves $O(1/\sqrt{T})$ convergence rate with almost $p*nw$ times less computation than A2C, where $p$ is the number of threads and $nw$ is the number of nodes. Hence, HSA2C achieves $p*nw$ iteration-speed-up. This iteration-speed-up converts to time-speed-up (real speed-up) provided that the communication between the nodes takes no time. The total communication time for HSA2C algorithms can be bounded as follows:
\begin{equation}
 T*Tc\leq TTc\leq M*T*Tc
\end{equation}
where $T$ is the number of iterations, $Tc$ is the communication time need for each master - worker exchange and $M$ is the global batch (see Alg.~\ref{alg51}). For simplicity, we assume that the $Tc$ is fixed and the same for all nodes. If the time needed for computing one update $Tu\leq Tc$, the total time needed by the distributed algorithm $DTT$ could be higher than that of the sequential A2C $STT$:
\begin{equation}
T*Tc+STT/M\leq DTT\leq M*T*Tc+STT/nw
\end{equation}
where $STT\leq T*Tc$, hence
\begin{equation}
    STT< STT(1+1/M)\leq DTT
\end{equation}

In such cases, exiting distributed algorithms like~\citep{adamski2018distributed} increase the local batch size so that $Tu$ increases resulting in lower stochastic gradient variance and allowing for higher learning rate to be used, hence better convergence rate. This introduces a trade-off between computational efficiency and sample efficiency. Increasing the batch size by a factor of $k$ increases the time needed for local computation  by $O(k)$ and reduces the variance proportionally to 1/k~\citep{bottou2018optimization}. Thus, a higher learning rate can be used. However, there is a limit on the size of the learning rate. This maximum learning speed can be improved on using HSA2C (Alg.~\ref{alg51} and Alg.~\ref{alg52}) which performs $B$ times less communication steps. These improvements are empirically studied in the main paper.

\begin{algorithm}[ht]
   \caption{HSA2C-Master}\label{alg51}
\begin{algorithmic}[1]
       \STATE \textbf{initialise:} global network weights $\bm{\theta}$ and $\bm{\theta_v}$, batch size $M$, global learning rate
       \FOR{$t=0,1,2,...T-1$}
        \STATE Collect $M$ updating vectors $(\bm w_1,\bm w_{v1}),...,(\bm w_M,\bm w_{vM})$ from the workers.
      \STATE Update the current estimate of the global parameters $(\bm{\theta},\bm{\theta_v})$ using $\sum_{m=1}^M (\bm w_m,\bm w_{vm})$
     \ENDFOR
\end{algorithmic}
\end{algorithm}

\begin{algorithm}[ht]
   \caption{HSA2C-Worker}\label{alg52}
\begin{algorithmic}[1]
  \STATE \textbf{initialise:} number of iterations of per-worker loop $B$, number of threads $p$, thread step counter $t\leftarrow 0$, shared counter $T\leftarrow 0$, thread gradient accumulator $\bm{g}\leftarrow 0$ and $\bm{g}_v\leftarrow 0$, shared update vector accumulator $\bm{G}\leftarrow 0$ and $\bm{G}_v\leftarrow 0$, mini-batch size $m$, mini-batch counter $n$, local shared parameter vector $\bm \theta$ and $\bm \theta_v$, local learning rate.
  \STATE Synchronise initial parameters  $\bm{ u_v}\leftarrow \bm \theta_v$ and $\bm{ u}\leftarrow\bm \theta $
  \STATE Fork p threads
  \WHILE{(MasterIsRun)}
      \STATE $t_{start}=t$
      \STATE Get state $\bm{s_t}$
        \REPEAT
            \STATE Read current values of $\bm u$ , denoted as $\bm{\hat u}$, from the shared memory.
            \STATE Take action $\bm{a_t}$ according to policy $\pi(\bm{a}|\bm{s_t};\bm{\hat u})$
            \STATE  Receive new state $\bm{s_{t+1}}$ and reward $r_t$
            \STATE $t\leftarrow t+1$
            \UNTIL{terminal $\bm{s_t}$ or $t-t_{start}==t_{max}$}
            \STATE  Read current values of $\bm{u_v}$ , denoted as $\bm{\hat u_v}$, from the shared memory.
            \STATE R=$\begin{cases} 0 & \text{for terminal } \bm{s_t}\\  V(\bm{s_t},\bm{\hat u_v}) &\text{for non-terminal } \bm{s_t} \end{cases}$
            \FOR{$i\in\{t-1,...,t_{start}\}$}
            \STATE $R\leftarrow r_i+\gamma R$
            \STATE Accumulate gradients wrt $\bm{\hat u}$: $\bm{g}\leftarrow \bm{g}+\nabla_{\bm{\hat u}}\log \pi(\bm{a_i}|\bm{s_i};\bm{\hat{u}})(R-V(\bm{s_i;\bm{\hat{u}_v}}))$
            \STATE Accumulate gradients wrt $\bm{\hat u_v}$: $\bm{g_v}\leftarrow \bm{g_v}+\frac{\partial(R-V(\bm{s_i;\bm{\hat{u}_v}}))^2}{\partial \bm{\hat u_v}}$
            \ENDFOR
        \STATE Accumulate mini-batch gradients: $\bm{G}\leftarrow \bm{G}+\bm{g}$ and $\bm{G}_v\leftarrow\bm{G}_v+ \bm{g_v}$
    \STATE $\bm g\leftarrow 0$
      \STATE $\bm{g_v}\leftarrow 0$
              \STATE $n\leftarrow n+1$
         \IF{$n==m$}
         \STATE $n=0$
         \STATE Update the shared network weights $\bm{ u}$ using $\bm{G}$ and the local learning rate
      \STATE Update the shared network weights $\bm{ u_v}$ using $\bm{G_v}$ and the local learning rate
    \STATE$\bm{G}=0$ and $\bm{G}_v=0$
      \STATE Update local steps: $T\leftarrow T+1$
      \IF{$T==B$}
      \STATE $T=0$
       \STATE Push the update vectors $\bm{ u}-\bm \theta$ and $\bm{ u}_v-\bm \theta_v$ to the master
       \STATE Pull global parameters $\bm \theta$ and  $\bm{\theta_v}$ from the master
       \STATE Synchronise initial parameters  $\bm{ u_v}\leftarrow \bm \theta_v$ and $\bm{ u}\leftarrow\bm \theta $
        \ENDIF
      \ENDIF
     \ENDWHILE
\end{algorithmic}
\end{algorithm}

\end{document}